\documentclass{article}
\usepackage{iclr2027_conference,times} 

\usepackage{hyperref}
\hypersetup{colorlinks=true,linkcolor=blue,citecolor=blue,urlcolor=blue}
\usepackage{url}
\usepackage{amsmath,amssymb,mathtools,amsthm}
\usepackage{bm}
\usepackage{graphicx}
\usepackage{xcolor}
\usepackage{booktabs,multirow,threeparttable,makecell}
\usepackage{algorithm}
\usepackage{algpseudocode}
\usepackage{enumitem}
\usepackage{caption}
\usepackage{subcaption}
\usepackage{wrapfig}
\usepackage{pifont}
\usepackage{microtype}
\usepackage{tikz}
\usetikzlibrary{arrows.meta,positioning,fit,shapes.geometric}

\newtheorem{proposition}{Proposition}
\newtheorem{theorem}{Theorem}

\newtheorem{corollary}[theorem]{Corollary}

\title{Dynamic Neural Koopman Distillation
for Fast Robot Control Using Diffusion Models}

\author{
Lei Zheng\thanks{Corresponding author:
\texttt{zack.zheng@nus.edu.sg}.}\kern0.5em$^{1}$,
Peiqi Yu$^{2}$,
Zengqi Peng$^{3}$,
Changliu Liu$^{2}$,
and Armin Lederer$^{1}$\\
$^{1}$National University of Singapore \quad
$^{2}$Carnegie Mellon University\\
$^{3}$The Hong Kong University of Science and Technology  \\
\texttt{\{zack.zheng,armin.lederer\}@nus.edu.sg}\\
\texttt{\{peiqiy,cliu6\}@andrew.cmu.edu},
\texttt{zpeng940@connect.ust.hk}
}

\iclrfinalcopy
\makeatletter
\def\iclr@runtitle{Preprint}
\makeatother

\begin{document}
\maketitle

\begin{abstract}
Diffusion models excel at generating diverse, multimodal trajectories for robotic control, yet their iterative denoising process introduces latency that can constrain real-time closed-loop execution. To address this problem, we propose a Dynamic Neural Koopman (DNK) distillation framework, which distills multistep diffusion inference into a single forward pass conditioned on observations and sampled noise. Specifically, we introduce a Factorized Dynamic Koopman (FDK) layer that models the denoising process through a latent linear transition. The noisy trajectory input is lifted into a latent space, where the FDK layer parameterizes a factorized linear operator with condition-dependent eigenvalues that adapt the noisy-to-denoised transition to the current context. We evaluate DNK across robot-control benchmarks that span locomotion,  state-based, and image-based manipulation, comparing against accelerated generative-policy baselines. The results demonstrate that DNK achieves competitive
or improved closed-loop performance relative to accelerated generative-policy baselines, while substantially reducing inference latency relative to the strongest high-performing one-step
baselines. On long-horizon manipulation tasks, DNK achieves competitive performance with substantially fewer inference parameters than the one-step policy baselines.  Hardware experiments on a physical Kinova manipulator further demonstrate millisecond-scale inference and reduced task completion time. A project page is available at \url{https://fdkoopman.github.io/}.
\end{abstract} 
\section{Introduction}
\label{sec:intro} \vspace{-1mm}
Diffusion models have emerged as a promising approach for robotic motion generation and control~\citep{urain2025survey,peng2025diffusion}. By shifting policy learning from deterministic state-action mappings to stochastic trajectory generation, diffusion-based policies have shown strong performance across a broad range of continuous tasks, including locomotion \citep{pan2024model}, manipulation \citep{chi2025diffusion}, motion planning \citep{yang2026dualshield}, and exploration \citep{samavi2025sicnav,cao2025dare}. Unlike traditional unimodal policies, these models effectively generate diverse and context-appropriate plans, which makes them well-suited to robots operating in dynamic environments. 

Despite this high expressivity,
% deploying diffusion policies for nonlinear physical systems in closed-loop configurations is limited by inference latency. 
inference latency remains a major obstacle to deploying diffusion policies on nonlinear physical systems.
% From the perspective of dynamical systems, standard diffusion-based trajectory generation requires repeated integration of a highly nonlinear reverse process.  
% This is typically formulated as a reverse stochastic differential equation that progressively denoises pure noise into physically feasible trajectories~\citep{song2021scorebased, ho2020denoising}.
During inference, a diffusion policy generates each trajectory through multiple reverse denoising steps, where a learned network sequentially transforms an initial noise sample into a task-conditioned trajectory~\citep{song2021scorebased, ho2020denoising}. As a result, these sequential updates can lead to substantial per-decision latency. This delay is incompatible with the low-latency updates required for high-frequency closed-loop control~\citep{hwangbo2019learning}.

% Such computational overhead precludes the real-time execution required for high-frequency feedback control, where agile locomotion and dynamic manipulation typically demand update rates between 50 and 200 Hz~\citep{hwangbo2019learning}. 
 % Consequently, uncompressed generative planners remain largely impractical for direct hardware deployment. 

To alleviate this latency bottleneck, researchers have explored diffusion acceleration and distillation frameworks, such as Consistency Models \citep{song2023consistency, song2024improved} and Progressive Distillation \citep{salimans2022progressive}. These methods substantially reduce the number of network evaluations required for diffusion sampling, thereby lowering inference cost. Recent studies have extended this paradigm to robot control \citep{prasad2024consistency,wang2025onestep}. 
However, existing one-step approaches typically parameterize the student policy as a direct conditional generator without an explicit context-dependent latent transition. Such context dependence is particularly relevant to robot control, where the noisy-to-denoised mapping can vary substantially across observations, contact conditions, and motion regimes~\citep{huang25a}.

Koopman operator theory provides a principled way to introduce dynamical structure into this setting by lifting nonlinear dynamics into a linear latent space \citep{bevanda2021koopman,shi26:koopman}. In robotics,  \citet{bi2025imitation} combine deep Koopman models with diffusion-based planning to enable dynamics learning and control under limited action supervision. However, such architectures still rely on iterative upstream diffusion planners. To streamline the denoising process itself, recent efforts in image synthesis have proposed using a fixed global latent transition to compress diffusion inference into a single step \citep{berman2025onestep}.  However, directly translating this static distillation framework to robot control is challenging 
since robotic trajectories can vary substantially across states, contact conditions, and motion regimes.  {A single global latent transition does not explicitly adapt its latent evolution to the current control context, which may limit its flexibility across heterogeneous trajectory regimes.}

In this paper, we propose Dynamic Neural Koopman (DNK) distillation, a  Koopman-inspired framework that compresses multistep diffusion inference into a one-step, context-conditioned latent transition for fast robot control.
Using Koopman-inspired latent dynamics~\citep{strasser2026overview, mezic2005spectral}, we approximate the multistep noisy-to-denoised transformation of the teacher through a context-conditioned factorized transition in a learned latent space. This design enables millisecond-level inference latency while supporting diverse candidate generation. The main contributions of this paper are summarized as follows:
\begin{enumerate}[itemsep=-2pt,topsep=-2pt,leftmargin=*]
    \item We present a one-step diffusion distillation framework for robot control by approximating the iterative reverse denoising process through a single linear transition in a learned latent space. 
    This largely reduces inference cost relative to multistep diffusion sampling, enabling high-frequency closed-loop control for nonlinear robotic systems.  

   \item  We introduce the Factorized Dynamic Koopman (FDK) layer, which parameterizes the latent transition using a shared modal basis and condition-dependent modal gains. Compared with a fixed global latent transition~\citep{berman2025onestep}, the FDK layer provides a more flexible parameterization of denoising dynamics through context-dependent spectral adaptation, with a spectral regularizer that discourages large modal gains. 
 
    \item  We evaluate the proposed method on D4RL locomotion and state-based and image-based manipulation benchmarks, and a physical Kinova Gen3 manipulator. Across these tasks, our approach maintains competitive teacher-level performance with substantially lower inference latency and a compact student architecture relative to high-capacity one-step baselines. Hardware experiments further demonstrate millisecond-scale inference, reduced task completion time, and obstacle-aware receding-horizon control. 
 \end{enumerate}

\section{Related Work}
\label{sec:related}\vspace{-2mm} 
\textbf{Diffusion Models for Robot Control:}  Recent work has increasingly applied diffusion models to robot control, where actions or trajectories are generated from a conditional distribution rather than as single deterministic outputs~\citep{ho2020denoising,urain2025survey}. This formulation is particularly useful when multiple feasible behaviors are compatible with the current context. Existing work has  applied  diffusion-based policies to manipulation \citep{chi2025diffusion}, locomotion \citep{pan2024model}, motion planning \citep{zhong2025coplanner,peng2025diffusion}, and exploration \citep{samavi2025sicnav,cao2025dare}. 
Unified software frameworks, such as CleanDiffuser, have facilitated more accessible, modular, and reproducible development of diffusion-based methods in robotics \citep{dong2024cleandiffuser}.
% Software frameworks, such as CleanDiffuser, have also made diffusion-policy training and benchmarking more reproducible in robotics~\citep{dong2024cleandiffuser}. 
Most of these methods generate actions through iterative denoising conditioned on the current observation or task context. While this procedure enables expressive conditional policies, it also entails substantial inference latency because multiple denoising steps are required before an action is generated. Such latency becomes a major limitation in fast closed-loop robot control, where actions must be updated at high frequency under tight latency constraints~\citep{urain2025survey}.
This motivates the development of faster diffusion-based policies that remain suitable for closed-loop control.

\textbf{Acceleration and Distillation of Diffusion Models:} Prior work on accelerating inference for diffusion models can be broadly grouped into two directions: fast samplers and distillation-based generators \citep{wang2025onestep,berman2025onestep}. Fast samplers reduce the number of denoising evaluations while retaining the original diffusion model.
Representative examples include Denoising Diffusion Implicit Models (DDIM)~\citep{song2021denoising}, DPM-Solver~\citep{lu2022dpm}, and EDM-style fast-sampling formulations~\citep{karras2022elucidating}. These methods improve efficiency, but they remain iterative and may be too slow for high-frequency closed-loop control.

Distillation-based methods instead train a student model to reproduce the teacher with fewer denoising steps, or in a single forward pass.  Progressive Distillation \citep{salimans2022progressive} reduces the number of sampling steps through teacher-student compression, while Consistency Models support direct or near-direct generation through Consistency Training (CT) and Consistency Distillation (CD) \citep{song2023consistency, song2024improved}.
These approaches can provide larger reductions in inference cost than sampler acceleration, making them attractive for fast control. Several recent works have extended these distillation paradigms to robotics~\citep{prasad2024consistency,wang2025onestep}. Although their training objectives exploit diffusion states or teacher-generated samples, the resulting students are typically parameterized as direct conditional generators without an explicit context-dependent latent transition. This distinction is  relevant to robot control, where the noisy-to-denoised  mapping can vary across observations, contact conditions, and motion regimes. Our work instead introduces context dependence directly into a structured latent transition of the one-step student.

\textbf{Koopman-Based Models for Nonlinear Dynamics:}
Koopman operator theory provides a principled way to represent nonlinear dynamics through linear evolution of observables in a lifted space \citep{bevanda2021koopman,strasser2026overview}. This perspective has motivated a broad literature on data-driven prediction and control, including neural variants that jointly learn lifted representations and approximate latent dynamics \citep{bi2025imitation,li2025continual}. In robotics, Koopman-based models have been used for system identification, prediction, and feedback control, especially when nonlinear dynamics are difficult to model directly~\citep{shi26:koopman}.
 
Recently,
% the Koopman Distillation Model (KDM) applies this idea to diffusion generation, encoding noisy samples into a learned latent space and propagating them through a linear operator before decoding the clean output~\citep{berman2025onestep}.
Koopman ideas have been introduced into generative modeling. The Koopman Distillation Model (KDM) shows that diffusion denoising can be compressed into a single latent linear evolution step for image generation \citep{berman2025onestep}. This result suggests that Koopman-based distillation is a promising approach to fast generative inference. However, existing Koopman diffusion formulations are developed primarily for image synthesis and rely on a static global operator.
In robot control, {the required behavior depends strongly on the current
state and task context, and may involve changing contact modes.}  This motivates our FDK parameterization, which retains a shared modal basis while adapting the modal gains to the conditioning context.
 
%  This places
% context-dependent modulation directly within the latent
% transition, providing a structured parameterization for
% compact, one-step policy distillation.

% In contrast to hierarchical modeling of intermediate diffusion
% trajectories~\citep{bai2025hierarchical}, our formulation targets
% compact endpoint distillation for receding-horizon robot control.

 \section{Problem Statement and Preliminaries}
\label{sec:problem}\vspace{-2mm}
We consider trajectory generation for high-frequency closed-loop robot control. Let $\mathbf{s}_t \in \mathcal{S} \subseteq \mathbb{R}^{m}$ and $\mathbf{a}_t \in \mathcal{A} \subseteq \mathbb{R}^{n}$ denote the robot state and action at control step $t$, respectively. Over a prediction horizon $H$, we use $\boldsymbol{\tau}$ to denote a predicted trajectory. Depending on the task and policy parameterization, $\boldsymbol{\tau}$ may represent a state--action trajectory $(\mathbf{s}_1,\mathbf{a}_1,\ldots,\mathbf{s}_H,\mathbf{a}_H)$ or an action trajectory $(\mathbf{a}_1,\ldots,\mathbf{a}_H)$.
% We consider trajectory generation for high-frequency closed-loop robot control. Let $\mathbf{s}_k \in \mathcal{S} \subseteq \mathbb{R}^{m}$ and $\mathbf{a}_k \in \mathcal{A} \subseteq \mathbb{R}^{n}$ 
% denote the robot state and action at the discrete time step $k$, respectively. Over a prediction horizon $H$, 
% a future state-action trajectory $\boldsymbol{\tau}$ is defined as $\boldsymbol{\tau} = (\mathbf{s}_1,\mathbf{a}_1,\dots,\mathbf{s}_H,\mathbf{a}_H)$.
  At each control step, the planner models a conditional trajectory distribution $\pi(\boldsymbol{\tau}\mid\mathbf{c})$, where $\mathbf{c}$ denotes the information available at the current step, such as the current observation, observation history, and task specification.

Let \(T_{\mathrm{inf}}\) denote the inference time required to generate trajectory samples, and let \(T_{\mathrm{ctrl}}\) denote the control period. 
For high-frequency closed-loop control, the generated action must be available within the control period, i.e., \(T_{\mathrm{inf}}\leq T_{\mathrm{ctrl}}\). 
Our goal is to learn a one-step student generator that substantially reduces \(T_{\mathrm{inf}}\) relative to the teacher diffusion policy while approximating the teacher-conditioned
outputs and supporting stochastic candidate generation.

% Let \(T_{\mathrm{inf}}\) denote the inference time required to generate
% trajectory samples, and let \(T_{\mathrm{ctrl}}\) denote the control period.
% For high-frequency closed-loop control with per-cycle replanning, the generated
% action must be available within the control period, i.e.,
% \(T_{\mathrm{inf}}\leq T_{\mathrm{ctrl}}\).
% Our goal is to learn a one-step student generator that substantially reduces
% \(T_{\mathrm{inf}}\) relative to the teacher diffusion policy while approximating
% the teacher-conditioned outputs and supporting stochastic candidate generation.

\subsection{Denoising Diffusion Probabilistic Models}
\label{sec:prelim_diffusion}\vspace{-1mm}
Denoising diffusion probabilistic models (DDPMs) learn a target distribution by reversing a Markovian noise-addition process \citep{ho2020denoising,song2021scorebased}. Let \(\boldsymbol{\tau}_0 \sim q(\boldsymbol{\tau}_0\mid\mathbf{c})\) denote a  denoised trajectory target conditioned on \(\mathbf{c}\). The forward diffusion process adds Gaussian noise over \(N\) steps according to a variance schedule \(\{\beta_k\}_{k=1}^N\), with \(\beta_k\in(0,1)\):
\begin{equation}
q(\boldsymbol{\tau}_k \mid \boldsymbol{\tau}_{k-1})
=
\mathcal{N}\bigl(
\boldsymbol{\tau}_k;
\sqrt{1-\beta_k}\,\boldsymbol{\tau}_{k-1},
\beta_k \mathbf{I}
\bigr),
\end{equation}
where \(\mathbf I\) denotes the identity matrix. Equivalently, $\boldsymbol{\tau}_k
=
\sqrt{\bar{\alpha}_k}\,\boldsymbol{\tau}_0
+
\sqrt{1-\bar{\alpha}_k}\,\boldsymbol{\epsilon},$
% \begin{equation}
%     \boldsymbol{\tau}_k
% =
% \sqrt{\bar{\alpha}_k}\,\boldsymbol{\tau}_0
% +
% \sqrt{1-\bar{\alpha}_k}\,\boldsymbol{\epsilon},
% \end{equation} 
where \(\bar{\alpha}_k=\prod_{i=1}^k(1-\beta_i)\) and \(\boldsymbol{\epsilon}\sim\mathcal{N}(\mathbf{0},\mathbf{I})\). For sufficiently large \(N\), the terminal distribution approaches \(\mathcal{N}(\mathbf{0},\mathbf{I})\).
The reverse process sequentially denoises \(\boldsymbol{\tau}_N\) to recover \(\boldsymbol{\tau}_0\). The resulting conditional diffusion policy can be written as
\begin{equation}
\label{eq:diffusion_policy}
\hspace{-3mm}\pi_\theta(\boldsymbol{\tau}_0 \mid \mathbf{c})
:=
\int
\mathcal{N}(\boldsymbol{\tau}_N;\mathbf{0},\mathbf{I})
\prod_{k=1}^{N}
p_\theta(\boldsymbol{\tau}_{k-1}\mid \boldsymbol{\tau}_k,\mathbf{c})
\, d\boldsymbol{\tau}_{1:N}.
\end{equation}
Here, \(d\boldsymbol{\tau}_{1:N}\) denotes integration over the intermediate diffusion variables \(d\boldsymbol{\tau}_1 \cdots d\boldsymbol{\tau}_N\). The term \(p_\theta(\boldsymbol{\tau}_{k-1}\mid \boldsymbol{\tau}_k,\mathbf{c})\) denotes the learned reverse denoising transition, defined as
\[
p_\theta(\boldsymbol{\tau}_{k-1}\mid \boldsymbol{\tau}_k,\mathbf{c})
=
\mathcal{N}\bigl(
\boldsymbol{\tau}_{k-1};
\boldsymbol{\mu}_\theta(\boldsymbol{\tau}_k,\mathbf{c},k),
\boldsymbol{\Sigma}_\theta(\boldsymbol{\tau}_k,\mathbf{c},k)
\bigr).
\]
The denoising network \(\boldsymbol{\epsilon}_\theta(\boldsymbol{\tau}_k,\mathbf{c},k)\) is trained with the standard DDPM noise-prediction objective $\mathcal{L}_{\text{DDPM}}
=
\mathbb{E}_{\boldsymbol{\tau}_0,\boldsymbol{\epsilon},k}
\bigl[
\|\boldsymbol{\epsilon}
-
\boldsymbol{\epsilon}_\theta(\boldsymbol{\tau}_k,\mathbf{c},k)\|^2
\bigr]$, 
where \(k\) is sampled uniformly from \(\{1,\dots,N\}\). 
This objective teaches the network to approximate the reverse denoising process through noise prediction.

The conditional diffusion policy in \eqref{eq:diffusion_policy} reveals the main computational bottleneck: generating one trajectory sample requires \(N\) sequential network evaluations, which limits real-time deployment in high-frequency receding-horizon control.
\subsection{Koopman Operator Theory and Neural Koopman Models}
\label{sec:prelim_koopman}\vspace{-1mm}
Koopman operator theory represents nonlinear dynamics through a linear evolution of observables in a lifted space \citep{strasser2026overview,mezic2005spectral}. For a discrete-time nonlinear system \(\mathbf{x}^{+}=\Phi(\mathbf{x})\), where \(\mathbf{x}^{+}\) denotes the next state and \(\Phi\) is the nonlinear state-transition map, the Koopman operator acts on a scalar observable \(g\) as $(\mathcal{K}g)(\mathbf{x}) = g(\Phi(\mathbf{x}))$. 
Although \(\Phi\) is generally nonlinear, \(\mathcal{K}\) is linear on the space of observables. In practice, this infinite-dimensional operator is approximated in a finite-dimensional lifted space \citep{strasser2026overview}.

 \begin{figure*}[t]
\centering
\includegraphics[width=0.925\textwidth]{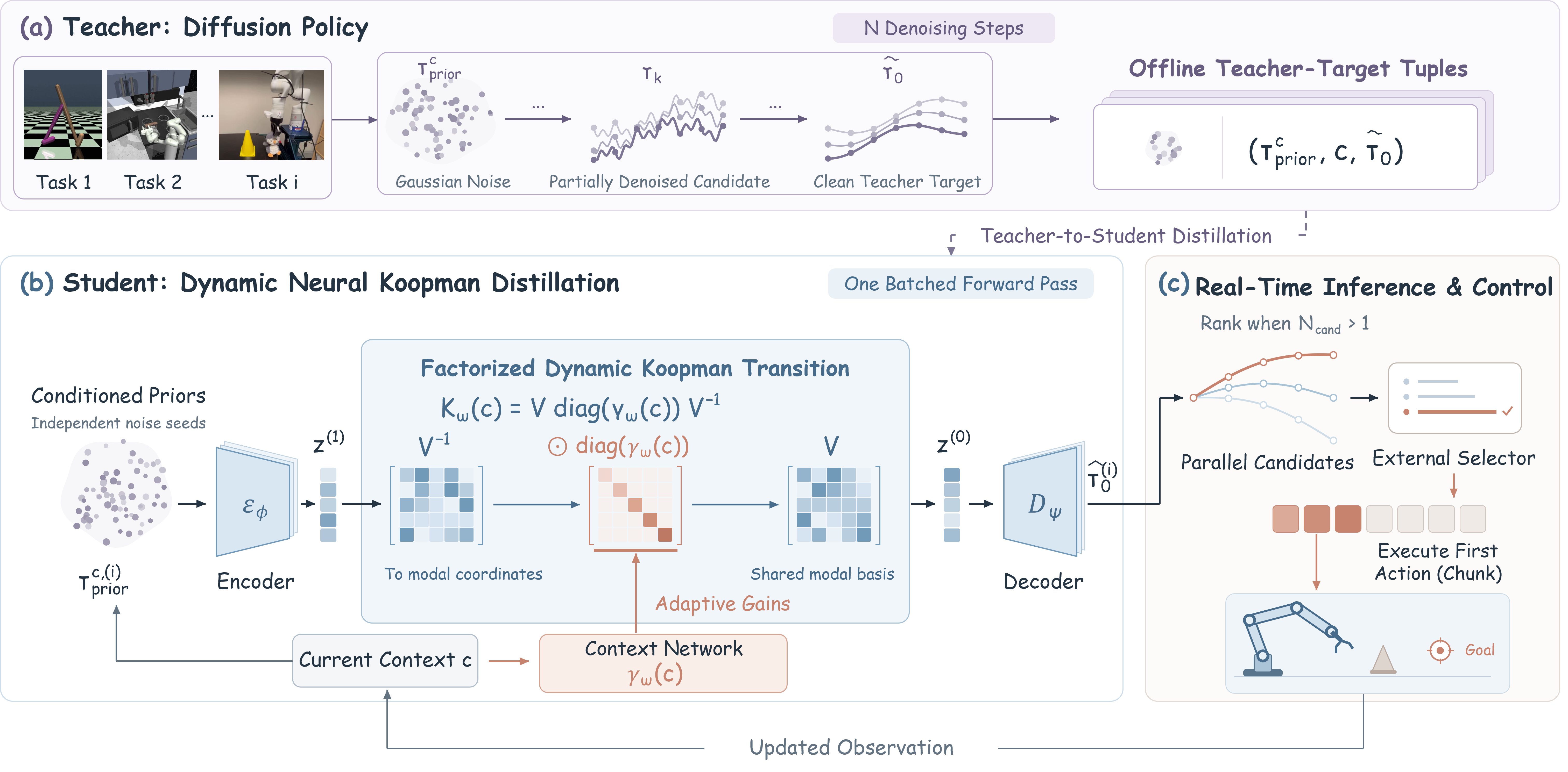}
\vspace{-2mm}
\caption{ Overview of the proposed DNK distillation framework.  A pretrained diffusion teacher generates offline teacher--target tuples
$(\boldsymbol{\tau}_{\mathrm{prior}}^{\mathbf c},
\mathbf c,\tilde{\boldsymbol{\tau}}_0)$ through multistep reverse diffusion.
The lightweight student encodes the conditioned prior samples and performs a single
FDK transition, 
using a shared modal basis and context-dependent latent spectrum. When multiple candidates are generated, the decoded sequences are ranked by an external selector and executed in closed loop with updated observations.
 }
\vspace{-4mm}
\label{fig:structure}
\end{figure*}

Neural Koopman models parameterize the observable lifting and reconstruction maps with neural networks and learn a latent space in which the dynamics are approximately linear. In this setting,
\begin{equation}
    \mathbf{z}_k=\mathcal{E}_{\phi}(\boldsymbol{\tau}_k), \qquad
\hat{\boldsymbol{\tau}}_k=\mathcal{D}_{\psi}(\mathbf{z}_k),
\end{equation}
where   \(\mathbf{z}_k\in \mathbb{R}^L\) denotes the lifted latent representation of the trajectory variable \(\boldsymbol{\tau}_k\), with \(L\) being the latent dimension; \(\mathcal{E}_{\phi}\) and \(\mathcal{D}_{\psi}\) denote the encoder and decoder with trainable parameters \(\phi\) and \(\psi\), respectively. Since diffusion inference proceeds from \(\boldsymbol{\tau}_N\) to \(\boldsymbol{\tau}_0\), we write the latent transition in the reverse direction. 
The learned finite-dimensional latent evolution is then approximated by 
\begin{equation}
    \mathbf{z}_{k} =\mathbf{A}\mathbf{z}_{k+1},
\end{equation} where \(\mathbf{A}\) denotes a finite-dimensional linear operator in the lifted latent space. 
 {For a time-homogeneous latent system, multi-step evolution can be represented compactly through powers of the same operator, i.e., $z_0 = A^k z_k$.}
%Armin: The core idea is that this linear representation lets you do multiple time steps by taking the exponent of A. Should we mention this here very briefly? 
As outlined by \cite{bevanda2023koopman}, under suitable assumptions, finite-dimensional lifted linear representations can approximate the original nonlinear map arbitrarily well.  This motivates our use of Koopman-inspired latent modeling to approximate  the multistep noisy-to-denoised transformation induced by diffusion inference.

 \section{Dynamic Neural Koopman Distillation Framework}
\label{sec:method} \vspace{-2mm}
This section presents the proposed DNK Distillation framework, as illustrated in Fig.~\ref{fig:structure}. 
The framework first constructs an offline teacher-target dataset by querying a pretrained DDPM teacher.
% with {mixed sampling temperatures for teacher-target generation}
It then trains a one-step student to map a task-conditioned Gaussian prior to a denoised target sequence through a context-conditioned lifted transition.  
The student is built on an FDK layer and trained using teacher-fidelity losses, control-oriented supervision, and structural regularization.
At test time, the distilled student is deployed in a receding-horizon control loop.

\subsection{Offline Teacher-Target Pair Generation}
\label{subsec:data_gen}    
Let $\pi_\theta$ denote the pretrained teacher diffusion policy.
We construct an offline distillation dataset by sampling conditioning
variables $\mathbf c$ from the dataset used to train the teacher.
For each context, we sample a Gaussian prior 
$\boldsymbol{\tau}_N \sim
\mathcal{N}(\mathbf{0},\sigma_N^2\mathbf{I})$
with the same horizon and per-step feature layout as the teacher
output, where $\sigma_N$ follows the sampling configuration of the teacher. For state--action trajectories, we obtain conditioned prior samples $\boldsymbol{\tau}_{\mathrm{prior}}^{\mathbf c}$ 
by fixing the observed state entries according to the teacher's
conditioning scheme. For action-only policies,
$\boldsymbol{\tau}_{\mathrm{prior}}^{\mathbf c}
=\boldsymbol{\tau}_N$, while the observation history, or its encoded
representation for high-dimensional observations, is provided separately
as conditioning to the denoising network.

Starting from $\boldsymbol{\tau}_{\mathrm{prior}}^{\mathbf c}$,
the teacher generates the target through multistep reverse diffusion:
\begin{equation}
\tilde{\boldsymbol{\tau}}_0
\sim
\pi_\theta\!\left(
\cdot \mid \mathbf{c};
\boldsymbol{\tau}_{\mathrm{prior}}^{\mathbf c}
\right),
\end{equation}
where the semicolon specifies the initial trajectory.
% indicates that the reverse diffusion process is initialized from \(\boldsymbol{\tau}_{\mathrm{prior}}^{\mathbf c}\).  
 Each distillation sample therefore takes the form \( (\boldsymbol{\tau}_{\mathrm{prior}}^{\mathbf c}, \mathbf c, \tilde{\boldsymbol{\tau}}_0) \),
with $\mathbf c$ retained for conditioning the student.

% Following this, the teacher performs multistep reverse diffusion to generate the corresponding denoised trajectory target:
% \begin{equation}
%     \tilde{\boldsymbol{\tau}}_0
%     \sim
%     \pi_\theta\!\left(
%     \cdot \mid \mathbf{c};
%     \boldsymbol{\tau}_{\mathrm{prior}}^{\mathbf c}
%     \right),
% \end{equation}
% where the semicolon indicates that the reverse diffusion process is initialized from \(\boldsymbol{\tau}_{\mathrm{prior}}^{\mathbf c}\). Each distillation sample therefore takes the form \( (\boldsymbol{\tau}_{\mathrm{prior}}^{\mathbf c}, \mathbf c, \tilde{\boldsymbol{\tau}}_0) \), where \(\mathbf c\) conditions   the teacher policy and is retained as part
% of the training tuple for the student. 

% where the semicolon indicates that the teacher reverse process is initialized from the conditioned prior \(\boldsymbol{\tau}_{\mathrm{prior}}^{\mathbf c}\).  
% Hence, each distillation example takes the form 
% \((\boldsymbol{\tau}_{\mathrm{prior}}^{\mathbf c}, \mathbf{c}, \tilde{\boldsymbol{\tau}}_0)\), where \(\mathbf c\) provides the conditioning input to both the teacher and the context-conditioned latent operator.

\subsection{Context-Conditioned Latent Transition via the FDK Layer}\label{subsec:fdk_arch} 
We propose an FDK layer to model one-step denoising as a context-dependent transition in a shared lifted latent space. 
% Let \(\boldsymbol{\tau}_{\mathrm{prior}} \in \mathbb{R}^{H\times d}\) denote the conditioned Gaussian prior trajectory, where $d = m+n$. Conditioning is injected through this prior representation by fixing the task-relevant entries of the prior tensor before encoding.
The student first maps the conditioned prior samples to a latent state
\begin{equation}
    \mathbf{z}^{(1)}
    =
\mathcal{E}_{\phi}\!\left(\boldsymbol{\tau}_{\mathrm{prior}}^{\mathbf{c}}\right)
    \in \mathbb{R}^{L},
\end{equation}
where \(L\) is the lifted latent dimension. The same encoder is applied to conditioned prior samples and denoised teacher targets, providing a shared latent representation for distillation.

 A static Koopman distillation model~\citep{berman2025onestep} applies a single global linear operator to all latent samples. This assumption can be restrictive for robotic state or action trajectory generation because the mapping from a noisy prior to a denoised trajectory may vary across observations, motion regimes, and task specifications.  To address this issue, we introduce an FDK layer that replaces the
fixed global transition with a factorized, context-conditioned family of latent linear operators.
Rather than predicting a dense \(L\times L\) operator independently for each
condition, FDK uses a shared learned modal basis together with condition-dependent
latent eigenvalues:
\begin{equation}\vspace{-0.2mm}
\mathbf z^{(0)}
=
\mathbf V
\left(
\mathbf V^{-1}\mathbf z^{(1)}
\odot
\boldsymbol{\gamma}_{\omega}(\mathbf c)
\right),
\label{eq:transition}\vspace{-1mm}
\end{equation}
where \(\odot\) denotes elementwise multiplication. 
The matrix
\(\mathbf V\in\mathbb R^{L\times L}\) denotes a shared nonsingular
modal basis, while \(\mathbf V^{-1}\) maps the lifted representation
\(\mathbf z^{(1)}\) into the corresponding modal coordinates.
The network
\(\boldsymbol{\gamma}_{\omega}:\mathbb R^{d_c}\rightarrow\mathbb R^{L}\)
maps the conditioning representation \(\mathbf c\) to a set of
condition-dependent latent eigenvalues, where \(d_c\) denotes the
dimension of the conditioning representation.
These coefficients modulate the latent modes according to the
information available at the current control step, such as the current
observation and task specification.
The transformed modal coordinates are then mapped back to the shared
latent space through \(\mathbf V\). 
Equivalently, the FDK layer defines
\begin{equation}
\mathbf z^{(0)}
=
\mathbf K_{\omega}(\mathbf c)\mathbf z^{(1)},
\qquad 
\mathbf K_{\omega}(\mathbf c)
=
\mathbf V
\operatorname{diag}\!\bigl(
\boldsymbol{\gamma}_{\omega}(\mathbf c)
\bigr)
\mathbf V^{-1}. 
\label{eq:fdk_operator}
\end{equation}
The context-dependent gains modulate the shared latent modes, providing
a structured form of adaptation across conditioning contexts. For any fixed condition $\mathbf c$, $\mathbf K_\omega(\mathbf c)$
defines a linear transition in the lifted state. Since
$\mathbf K_\omega(\mathbf c)$ is similar to
$\operatorname{diag}(\boldsymbol{\gamma}_\omega(\mathbf c))$,
its eigenvalues are exactly $\gamma_{\omega,\ell}(\mathbf c)$, while
the columns of $\mathbf V$ define the shared modal basis. 
Thus, FDK adapts the spectrum of the latent transition to the current
context while retaining a shared modal basis.
This structure introduces context dependence without predicting a dense
\(L\times L\) operator for each context: only the \(L\) modal gains
\(\boldsymbol{\gamma}_{\omega}(\mathbf c)\) vary with the conditioning
context, whereas the modal basis $\mathbf V$ is shared.
Consequently, the condition-dependent degrees of freedom scale as
\(L\), rather than \(L^2\) for an unrestricted context-specific dense
transition.
Compared with the fixed global transition used in static Koopman
distillation~\citep{berman2025onestep}, FDK provides structured
context adaptation while preserving a shared latent representation.

For a fixed latent representation and shared modal basis, the optimal
context-dependent modal gains attain no larger population
latent-prediction risk than the best constant gain vector, with strict
improvement whenever the optimal gains vary with the conditioning
context. Appendix~\ref{app:conditional_gain_proof} formalizes this result.
After the FDK transition, the decoder reconstructs the predicted
denoised trajectory:
$\hat{\boldsymbol{\tau}}_0=
\mathcal{D}_{\psi}\bigl(\mathbf{z}^{(0)}\bigr)$.

\subsection{Training Objective}
\label{subsec:loss}
Given an offline teacher-target sample \((\boldsymbol{\tau}_{\mathrm{prior}}^{\mathbf c,(i)}, \tilde{\boldsymbol{\tau}}_0^{(i)})\) from the dataset defined in Section~\ref{subsec:data_gen}, the student predicts a denoised trajectory \(\hat{\boldsymbol{\tau}}_0^{(i)}\). 
We propose a training objective that aligns the student with the teacher while emphasizing control-relevant predictions and regularizing the latent transition.
% The training objective combines teacher-fidelity losses, control-oriented supervision, structural regularization, and teacher-quality reweighting.

% The training objective combines teacher-fidelity losses, control-oriented supervision, structural regularization, and a sample reweighting scheme based on teacher-side quality scores.

% \subsubsection{Teacher-Fidelity Losses}  
\textbf{Teacher-Fidelity Losses:} To distill the teacher into a one-step student, we align the student with the teacher using  denoised-trajectory reconstruction, {latent-target alignment}, and trajectory prediction: 
\begin{align}
\mathcal{L}_{\mathrm{rec}}
&=
\mathbb{E}\!\left[
\|\mathcal{D}_{\psi}(\mathcal{E}_{\phi}(\tilde{\boldsymbol{\tau}}_0))-\tilde{\boldsymbol{\tau}}_0\|_2^2
\right] , \label{eq:loss_rec} \\
\mathcal{L}_{\mathrm{lat}}
&=
% \mathbb{E}\!\left[
% \|\mathbf{z}^{(0)}-\operatorname{sg}(\mathcal{E}_{\phi}(\tilde{\boldsymbol{\tau}}_0))\|_2^2
% \right] =
\mathbb{E}\!\left[
\left\|\mathbf z^{(0)}
-
\operatorname{sg}
\!\left(
\mathcal{E}_{\phi}(\tilde{\boldsymbol{\tau}}_0)
\right)
\right\|_2^2
\right], \label{eq:loss_lat}\\
\mathcal{L}_{\mathrm{pred}} 
&=
\mathbb{E}\!\left[
\|\mathcal{D}_{\psi}(\boldsymbol{z}^{(0)}
)-\tilde{\boldsymbol{\tau}}_0\|_2^2
\right]. \label{eq:loss_pred}
\end{align}
Here, \(\mathcal{L}_{\mathrm{rec}}\)  regularizes the encoder--decoder pair so that denoised teacher trajectories are represented in a shared latent space;   
$\mathcal L_{\mathrm{lat}}$ aligns the predicted denoised latent
$\mathbf z^{(0)}= 
\mathbf K_{\omega}(\mathbf c)
\mathcal{E}_{\phi}
\!\left(
\boldsymbol{\tau}_{\mathrm{prior}}^{\mathbf c}
\right)$ with the encoded teacher target, providing direct supervision for the output of the context-conditioned
FDK transition. The stop-gradient operator $\operatorname{sg}(\cdot)$ detaches the
encoded teacher target, providing a latent target for the FDK-evolved
state $\mathbf z^{(0)}$ at each optimization step.
Finally, $\mathcal L_{\mathrm{pred}}$ directly supervises the decoded
trajectory to match the teacher target.

% The stop-gradient operator $\operatorname{sg}(\cdot)$ blocks gradients
% through the encoded teacher-target branch, so
% $\mathcal{E}_{\phi}(\tilde{\boldsymbol{\tau}}_0)$ serves as a detached latent target for the FDK-evolved state
% $\mathbf{z}^{(0)}$ during the current optimization step. 
% Finally, \(\mathcal{L}_{\mathrm{pred}}\) directly supervises the decoded trajectory to match the teacher target.

% For action-centric manipulation (Push-T, Relay-Kitchen), we disable the classical Koopman autoencoder terms \(\mathcal{L}_{\mathrm{rec}}\) and \(\mathcal{L}_{\mathrm{pred}}\). Packed trajectories contain zero-filled future-state slots that are not valid teacher targets, and closed-loop performance depends on executable actions rather than full-state reconstruction. We therefore supervise primarily with \(\mathcal{L}_{\mathrm{act}}\), and retain latent alignment and spectral regularization (\(\mathcal{L}_{\mathrm{lat}}\), \(\mathcal{L}_{\mathrm{spec}}\)) together with the FDK operator.

% \subsubsection{Control-Oriented Supervision}
\textbf{Control-Oriented Supervision:}
Since only the first action or action chunk is executed in a receding-horizon manner, control performance depends strongly on near-term action prediction. We therefore augment the objective with a weighted action-sequence loss defined directly on the predicted and teacher-target trajectories. Let
\(\Pi_a(\boldsymbol{\tau})=[\mathbf a_1^\top,\dots,\mathbf a_H^\top]^\top \in \mathbb{R}^{Hn}\)
denote the action sequence associated with \(\boldsymbol{\tau}\). We define
\begin{equation}\vspace{-1mm}
\mathcal{L}_{\mathrm{act}}
=
\mathbb{E}\!\left[
\left\|
\Pi_a(\hat{\boldsymbol{\tau}}_0)
-
\Pi_a(\tilde{\boldsymbol{\tau}}_0)
\right\|_{\mathbf W_a}^{2}
\right],
\label{eq:loss_act}\vspace{-0.2mm}
\end{equation}
where \(\|\mathbf x\|_{\mathbf W_a}^{2}=\mathbf x^\top \mathbf W_a \mathbf x\), and \(\mathbf W_a\) denotes a diagonal action-weighting matrix. 

%  \textcolor{blue}{For action-centric manipulation tasks such as Push-T, we set
% $\alpha_{\mathrm{rec}}=\alpha_{\mathrm{pred}}=0$.
% Prediction supervision is retained through
% $\mathcal L_{\mathrm{act}}$ on the decoded action sequence;
% Packed trajectories contain zero-filled future-state slots that are not valid teacher targets, and closed-loop performance depends on executable actions rather than full-state reconstruction. These experiments use a scale-stabilized latent objective
% in place of the basic squared-error alignment above,
% together with the applicable structural regularizers.}

\textbf{Structural Regularization:}
To regularize the factorized latent transition, we introduce a structural regularizer that aims at limiting excessive modal amplification:
\begin{equation}\vspace{-1mm}
\mathcal{L}_{\mathrm{spec}}
=
\mathbb{E}\!\left[
\frac{1}{L}\sum_{\ell=1}^{L}\max\bigl(0, |\gamma_{\omega,\ell}(\mathbf c)|-1\bigr)
\right],
\label{eq:spec}\vspace{-0.2mm}
\end{equation}
where \(\gamma_{\omega,\ell}(\mathbf c)\) denotes the \(\ell\)-th  coefficient produced by the FDK modulation network. This term penalizes modal eigenvalue magnitudes that exceed unity.  The corresponding spectral properties and the role of the
modal-basis conditioning are discussed in
Appendix~\ref{app:fdk_spectral}.
 
\textbf{Total Objective:} Thus, the final training objective is 
\begin{align}
\mathcal{L}
=&\;
\alpha_{\mathrm{rec}}\mathcal{L}_{\mathrm{rec}}
+\alpha_{\mathrm{lat}}\mathcal{L}_{\mathrm{lat}}
+\alpha_{\mathrm{pred}}\mathcal{L}_{\mathrm{pred}}  
+ \alpha_{\mathrm{spec}}\mathcal{L}_{\mathrm{spec}} +\mathcal{L}_{\mathrm{act}},
\label{eq:total_loss}
\end{align} 
 where \(\alpha_{\mathrm{rec}}\), \(\alpha_{\mathrm{lat}}\), \(\alpha_{\mathrm{pred}}\), and \(\alpha_{\mathrm{spec}}\)  are the corresponding loss weights.  The active loss terms and their benchmark-specific weights are summarized
in Appendix~\ref{subsec:sim_setup}. 
% in action-only manipulation tasks,
% full-trajectory reconstruction and prediction losses are omitted as
% only action targets are available.

% During training, we maintain and save an exponential moving average checkpoint of the student parameters.

\subsection{Inference: Batched One-Step Receding-Horizon Control}
\label{subsec:inference}
\vspace{-2mm}
After training, the distilled student is deployed in a receding-horizon control manner, as illustrated in Fig.~\ref{fig:structure}. At each control step, a conditioned Gaussian prior trajectory \(\boldsymbol{\tau}_{\mathrm{prior}}\) is constructed by sampling Gaussian noise and encoding current task-relevant context into the prior representation.
% In the MuJoCo setting, this is implemented by fixing the first observation component to the current normalized state, while in the Kinova setting, the conditioned input additionally includes task-specific robot observations.

When multi-candidate generation is used, we draw \(N_{\mathrm{cand}}\) independent noise seeds and evaluate the student on the resulting batch of conditioned prior samples. This yields a set of candidate trajectories in parallel.  These candidates are then ranked by an external selector, such as an Implicit Q-Learning (IQL) critic with argmax selection~\citep{kostrikov2022offline} or a   classifier~\citep{dong2024cleandiffuser}, which assigns each candidate a task-dependent control score. The controller executes the first action of the highest-scoring candidate and replans at the next control step using updated observations.
This inference scheme separates stochastic candidate generation from control-oriented candidate selection: independent noise seeds provide multiple candidate trajectories, while the external selector favors candidates with higher task-dependent control scores~\citep{pmlr-v283-huang25a}.    

\section{Experiments}
\label{sec:sim} 
In this section, we evaluate the proposed DNK distillation framework on D4RL MuJoCo locomotion, state-based and image-based Push-T, long-horizon Franka manipulation benchmarks, and a physical Kinova Gen3  control task.
\subsection{Simulation Setup}
\label{sec:setup}\vspace{-1mm}

\textbf{Benchmarks and Teacher:}
We evaluate the proposed method on the following benchmarks. The detailed experimental setup is given in Appendix~\ref{app:experimental_setup}.
\begin{enumerate} [itemsep=-2pt,topsep=-2pt,leftmargin=*]
    \item \textbf{D4RL locomotion}~\citep{fu2020d4rl}: we consider  halfcheetah  and
  walker2d (17D observations, 6D actions) tasks. They are useful for testing the performance under strongly nonlinear dynamics and rapidly changing motion regimes, where the required behavior varies substantially across states. 
    
    \item \textbf{Push-T}~\citep{florence2022implicit,chi2025diffusion}:  a contact-rich planar manipulation task in which a circular end-effector
    pushes a T-shaped block to a fixed target pose. We evaluate both 5D state
    and high-dimensional $96\times96$ RGB observations to examine contact-rich trajectory generation.

    \item \textbf{Franka Kitchen}~\citep{gupta2019relay}: a long-horizon manipulation benchmark with seven kitchen subtasks, where performance is measured by completion of up to four subtasks using 60D state observations and 9D actions. It is useful for evaluating whether the distilled policy can maintain effective behavior over long horizons and across changing task stages.

% a long-horizon manipulation benchmark with seven available kitchen subtasks, where performance is summarized by completion of up to four distinct subtasks using 60D state observations and 9D actions.
    
    % \item \textbf{Robomimic}~\citep{mandlekar2021what}: The single-arm proficient-human tasks Lift, Can and Square. These tasks provide manipulation
    % settings with increasing contact and geometric complexity and test generalization across tasks and observation modalities.
\end{enumerate}

\textbf{Student Models and Baselines:}
We compare the proposed method against three groups of baselines. First, we include the diffusion teacher: DDPM~\citep{ho2020denoising} and EDM~\citep{karras2022elucidating} as high-quality references using the official CleanDiffuser~\citep{dong2024cleandiffuser} implementation. Second, we compare with two static Koopman baselines, KDM and KDM-F \citep{berman2025onestep}. 
Third, we compare with one-step diffusion policies, OneDP-S~\citep{wang2025onestep}, and consistency-based baselines, including CT~\citep{song2024improved},  CD~\citep{song2023consistency}, and Consistency Policy (CP)~\citep{prasad2024consistency}.

\begin{wraptable}{r}{0.6\textwidth}
\vspace{-5mm}
\centering
\captionsetup{type=table}
\caption{Quantitative results on D4RL locomotion tasks. $T_D$ and $T_E$ denote task-specific DDPM and EDM teachers, respectively. NFE denotes the number of sampling network evaluations per generated action chunk.}
\label{tab:sim_locomotion_results}
\vspace{-3mm}

\scriptsize
\renewcommand{\arraystretch}{1.03}
\setlength{\tabcolsep}{2.2pt}

\resizebox{\linewidth}{!}{%
\begin{tabular}{@{}l c *{2}{c} *{2}{c}@{}}
\toprule
& & \multicolumn{2}{c}{\textbf{HalfCheetah}} & \multicolumn{2}{c}{\textbf{Walker2D}} \\
\cmidrule(lr){3-4}\cmidrule(lr){5-6}
Method & NFE
& Score~$\uparrow$ &Latency~(ms)~$\downarrow$ 
& Score~$\uparrow$ &Latency~(ms)~$\downarrow$ \\
\midrule
DDPM Teacher~$T_D$~\citep{dong2024cleandiffuser} & 20
& $0.867\pm0.004$ & $36.94\pm0.63$
& $1.109\pm 0.000$ &  $49.65\pm0.12$  \\
EDM Teacher~$T_E$~\citep{dong2024cleandiffuser} & $18$
& $0.842\pm0.012$ & $6.72\pm0.02$ 
& $1.108\pm0.001$ & $6.67\pm0.01$ \\
\midrule
CD~\citep{song2023consistency} & $1$
& $0.428\pm0.013$ & $1.08\pm2.32$
& $0.689\pm0.010$ & $1.09\pm2.33$  \\
CT~\citep{song2024improved} & $1$
& $0.403\pm0.025$ & $1.08\pm2.31$ 
& $0.387\pm0.017$ & $1.09\pm2.37$ \\
CP~\citep{prasad2024consistency} ($T_E$) & $1$
& $0.676\pm0.012$ & $0.981\pm0.01$ 
& $1.059\pm0.013$ & $0.99\pm0.01$ \\
OneDP-S~\citep{wang2025onestep} ($T_D$) & $1$
& $0.748\pm0.030$ & $2.03\pm0.11$ 
& $1.105\pm0.000$ & $2.03\pm0.11$ \\
KDM~\citep{berman2025onestep} ($T_D$) & $1$
& $0.553\pm0.024$ & $0.34\pm0.03$
& $0.620\pm0.011$ & $0.35\pm0.04$ \\
KDM-F~\citep{berman2025onestep} ($T_D$) & $1$
& $0.510\pm0.004$ & $0.86\pm0.03$ 
& $0.396\pm0.066$ & $0.87\pm0.03$ \\
\textbf{DNK (Ours)} ($T_D$) & $1$
& $\mathbf{0.873}\pm\mathbf{0.007}$ & $\mathbf{0.68}\pm\mathbf{0.39}$
& $\mathbf{1.095}\pm\mathbf{0.001}$ & $\mathbf{0.60}\pm\mathbf{0.42}$\\
\bottomrule
\end{tabular}%
}
\vspace{-4mm}
\end{wraptable} 
\textbf{Evaluation Protocol:} Each method is evaluated over three evaluation groups of
50 rollouts, using identical environment initializations and evaluation
settings across methods. Benchmark-specific details, such as rollout horizons and candidate budgets,  are given in Appendix~\ref{app:experimental_setup}. We report closed-loop task performance, inference latency, and model size. Locomotion performance is the normalized return and manipulation performance is the corresponding closed-loop metric (normalized task score or success rate), both summarized as mean~$\pm$~standard deviation across seeds. Latency is the per-step policy decision time on an NVIDIA RTX~5090 GPU with CUDA synchronization, and model size is the number of parameters.

\subsection{Simulation Results} 
\begin{wrapfigure}{r}{0.45\textwidth}
    \vspace{-4mm}
    \centering
    \captionsetup{font=small}
    \includegraphics[width=0.98\linewidth]
    {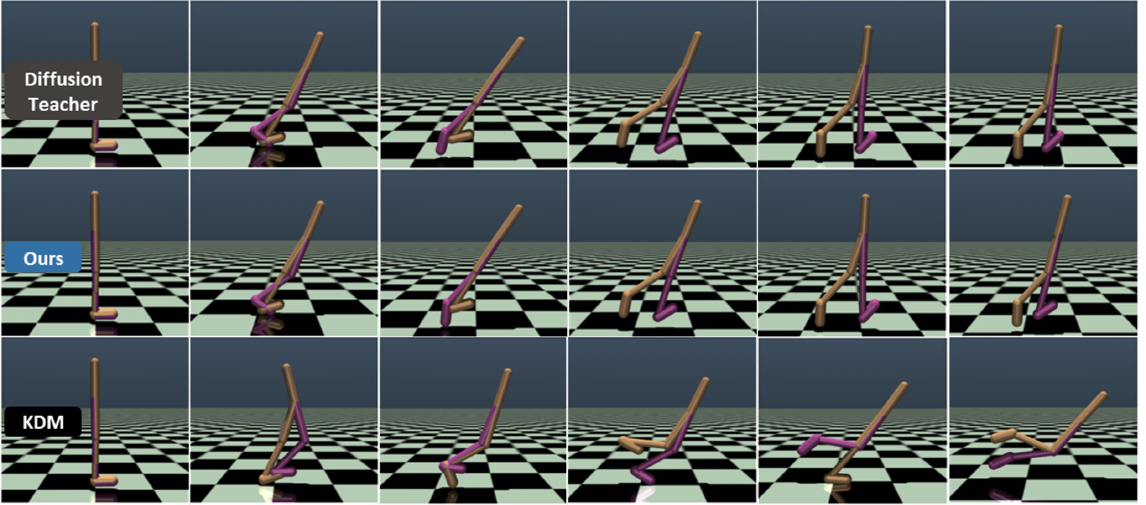}
    \vspace{-2mm}
    \caption{Representative Walker2d rollouts. The static KDM~\citep{berman2025onestep} fails to maintain a stable gait,
    whereas the diffusion teacher and DNK maintain coordinated forward motion.}
    \label{fig:kdm_failure}  \vspace{-4mm}
\end{wrapfigure}

\textbf{Locomotion.}
Table~\ref{tab:sim_locomotion_results} summarizes the quantitative
results in the agile locomotion benchmarks. DNK achieves competitive normalized
returns on both tasks, reaching \(0.873 \pm 0.007\) on HalfCheetah and
\(1.095 \pm 0.001\) on Walker2d.  {
Compared with OneDP-S, DNK achieves a higher return on HalfCheetah
while requiring substantially lower inference latency.
On Walker2d, OneDP-S attains a slightly higher mean return
(\(1.105\) vs.\ \(1.095\)), whereas DNK reduces latency from
\(2.03\) to \(0.60\)\,ms, corresponding to a \(3.38\times\) speedup.
DNK also achieves higher returns and lower latency than CP on both tasks.} Although the static Koopman baselines KDM and KDM-F maintain low inference latency, they achieve substantially lower returns than DNK on both environments. As illustrated in Fig.~\ref{fig:kdm_failure}, the static KDM policy fails to maintain a stable forward gait in the representative Walker2d rollout. 
Relative to the 20-step DDPM teacher, DNK reduces 
latency by \(54.3\times\) on HalfCheetah and \(82.8\times\) on Walker2d while retaining almost identical performance. 
\vspace{-0.5mm} 
\begin{table*}[t]
\centering
\caption {Closed-loop performance comparison of different approaches on Push-T. NFE denotes the number of sampling network evaluations per generated action chunk. Params. counts deployed neural-network parameters in millions, including frozen visual encoders.
Image DNK includes its $2.56$M action network and $11.37$M frozen visual encoder.
}
\label{tab:pusht_score_latency}\vspace{-0mm}
\label{tab:pusht_robomimic_score_latency}
\footnotesize
\setlength{\tabcolsep}{4pt}
\renewcommand{\arraystretch}{1.08}
\resizebox{\textwidth}{!}{%
\begin{tabular}{@{}l c *{3}{c} *{3}{c}@{}}
\toprule
& & \multicolumn{3}{c}{\textbf{Push-T State}} & \multicolumn{3}{c}{\textbf{Push-T Image + proprioception}} \\
\cmidrule(lr){3-5}\cmidrule(lr){6-8}
Method & NFE
& Score~$\uparrow$ &Latency~(ms)~$\downarrow$ & Params.~(M)~$\downarrow$
& Score~$\uparrow$ &Latency~(ms)~$\downarrow$ & Params.~(M)~$\downarrow$ \\
\midrule
DDPM Teacher~$T_D$~\citep{dong2024cleandiffuser}
& $5$
& $0.937{\pm}0.047$ & $30.592{\pm}0.077$ & $68.88$
& $0.845{\pm}0.027$ & $31.554{\pm}0.073$ & $80.39$ \\
EDM Teacher~$T_E$~\citep{dong2024cleandiffuser}
& $158$
& $\mathbf{0.945{\pm}0.021}$ & $1802.309{\pm}5.565$ & $248.01$
& $\mathbf{0.913{\pm}0.024}$ & $1809.775{\pm}4.711$ & $262.71$ \\
\midrule
CD~\citep{song2023consistency}
& $1$
& $0.859{\pm}0.007$ & $3.162{\pm}0.013$ & $68.88$
& $0.720{\pm}0.013$ & $4.768{\pm}0.037$ & $80.39$ \\
CT~\citep{song2024improved}
& $1$
& $0.887{\pm}0.069$ & $3.192{\pm}0.041$ & $68.88$
& $0.820{\pm}0.052$ & $4.680{\pm}0.055$ & $80.39$ \\
CP~\citep{prasad2024consistency} ($T_E$)
& $1$
& $0.927{\pm}0.017$ & $11.797{\pm}0.061$ & $251.68$
& $0.815{\pm}0.026$ & $12.922{\pm}0.027$ & $266.38$ \\
OneDP-S~\citep{wang2025onestep} ($T_D$)
& $1$
& $0.923{\pm}0.021$ & $6.162{\pm}0.009$ & $68.88$
& $0.825{\pm}0.013$ & $7.053{\pm}0.028$ & $80.39$ \\
KDM~\citep{berman2025onestep} ($T_D$)
& $1$
& $0.378{\pm}0.049$ & $0.618{\pm}0.006$ & $1.12$
& $0.515{\pm}0.056$ & $1.347{\pm}0.012$ & $12.75$ \\
KDM-F~\citep{berman2025onestep} ($T_D$)
& $1$
& $0.272{\pm}0.012$ & $0.903{\pm}0.003$ & $1.91$
& $0.519{\pm}0.074$ & $1.530{\pm}0.009$ & $13.54$ \\
\textbf{DNK (Ours)} ($T_D$)
& $1$
& $\mathbf{0.931{\pm}0.037}$ & $\mathbf{1.283{\pm}0.007}$ & $\mathbf{2.37}$
& $\mathbf{0.853{\pm}0.041}$ & $\mathbf{1.780{\pm}0.005}$ & $\mathbf{13.93}$ \\
\bottomrule
\end{tabular}%
}
\vspace{-3mm}
% \begin{minipage}{0.99\textwidth}
% \footnotesize
% \textit{Model size.}
% Params. counts deployed neural-network parameters in millions, including frozen visual encoders.
% Image DNK includes its $2.56$M action network and $11.37$M frozen visual encoder.
% \end{minipage}\vspace{-7mm}
\end{table*}

\textbf{Push-T.} 
% We next validate the performance of our approach in manipulation with both state and high-dimensional image-plus-proprioception
% observations.
Table~\ref{tab:pusht_score_latency}  shows that DNK achieves the highest mean score among students who take one step under both observation settings, with substantially lower inference latency than non-Koopman methods.
% reaching \(0.931\pm0.037\) with state observations and \(0.853\pm0.041\) with image observations. 
% These results are close to the corresponding DDPM teacher with a significantly lower latency. 
Notably, DNK obtains this competitive performance with a substantially smaller deployed model: the state-based student contains only \(2.37\)M parameters, compared with \(68.88\)M for OneDP-S and \(251.68\)M for CP, corresponding to approximately \(29\times\) and \(106\times\) reductions, respectively. 
For image-based control,  DNK reduces the
deployed parameter count by approximately \(5.8\times\) and
\(19.1\times\) compared to OneDP-S and  CP, respectively. Although the smaller KDM variants require fewer parameters and lower latency, their substantially lower task scores indicate a less favorable performance--latency--capacity trade-off. Overall, these results show that DNK compresses both the iterative sampling procedure and the deployed policy architecture without requiring the student to inherit the large teacher network.
\vspace{-0.5mm}
 \begin{wraptable}{r}{0.6\textwidth}
\vspace{-4.5mm}
\centering
\captionsetup{type=table}
\caption{
Quantitative comparison on Franka-Kitchen. 
NFE denotes the number of sampling-network evaluations per generated action chunk.
}
\label{tab:relay_kitchen_score_latency}  
\vspace{-2mm}

\scriptsize
\renewcommand{\arraystretch}{1.03}
\setlength{\tabcolsep}{2.2pt}

\resizebox{\linewidth}{!}{%
\begin{tabular}{@{}llcccc@{}}
\toprule
Task & Method & NFE
& Score~$\uparrow$
& Latency (ms)~$\downarrow$
& Params. (M)~$\downarrow$ \\
\midrule

\multirow{9}{*}{Franka Kitchen}
& DDPM Teacher $T_D$~\citep{dong2024cleandiffuser}
& 50
& $0.985 \pm 0.000$
& $115.427 \pm 0.809$
& $68.92$ \\

& EDM Teacher $T_E$~\citep{dong2024cleandiffuser}
& 158
& $0.990 \pm 0.010$
&$ 681.062  \pm 0.494 $
& $251.19$ \\

\cmidrule(lr){2-6}

& CD~\citep{song2023consistency}
& 1
& $0.440 \pm 0.040$
& $3.690 \pm 0.067$
& $68.92$ \\

& CT~\citep{song2024improved}
& 1
& $0.803 \pm 0.028$
& $3.640 \pm 0.010$
& $68.92$ \\

& CP~\citep{prasad2024consistency} ($T_E$)
& 1
& $0.857 \pm 0.035$
& $12.026 \pm 0.041$
& $254.86$ \\

& OneDP-S~\citep{wang2025onestep} ($T_D$)
& 1
& $0.975 \pm 0.013$
& $6.434 \pm 0.045$
& $68.92$ \\

& KDM~\citep{berman2025onestep} ($T_D$)
& 1
& $0.462 \pm 0.050$
& $0.607 \pm 0.008$
& $1.78$ \\

& KDM-F~\citep{berman2025onestep} ($T_D$)
& 1
& $0.488 \pm 0.031$
& $0.818 \pm 0.003$
& $1.83$ \\ 

& \textbf{DNK (Ours)} ($T_D$)
& $\mathbf{1}$
& $\mathbf{0.973{\pm}0.016}$
& $\mathbf{1.598{\pm}0.004}$
& $\mathbf{2.46}$ \\

\bottomrule
\end{tabular}%
} 
\vspace{-3mm}
\end{wraptable}

 \textbf {Long-horizon Franka Kitchen.} 
% To further examine whether this compact generation and competitive performance extends to long-horizon manipulation. 
Table~\ref{tab:relay_kitchen_score_latency} summarizes the results on the Franka-kitchen benchmark. 
DNK achieves $0.973\pm0.016$, compared with
$0.975\pm0.013$ for OneDP-S and $0.985\pm 0.000$
for the DDPM teacher. Although OneDP-S attains a slightly higher mean score,
DNK uses \(2.46\)M parameters, {approximately \(28\times\) fewer than OneDP-S.}
Moreover, DNK achieves a mean inference latency of
\(1.598\)\,ms,  {corresponding to a \(4.03\times\)
speedup over OneDP-S and a \(7.53\times\) speedup over CP.}
 Additionally, DNK   achieves higher task scores than CD, CT, CP, KDM, and KDM-F.
These results demonstrate a favorable
performance--latency--capacity trade-off in long-horizon manipulation. % A detailed ablation regarding the performance and capacity size is provided in Appendix~\ref{app:relay_kitchen_capacity}
  
% Across these benchmarks, DNK combines single-step generation with a compact student architecture while retaining competitive closed-loop performance.  

\subsection{Ablation Study}  
\begin{wrapfigure}{r}{0.28\textwidth}
    \vspace{-3mm}
    \centering
    \captionsetup{font=small}
    \includegraphics[width=0.98\linewidth]
    {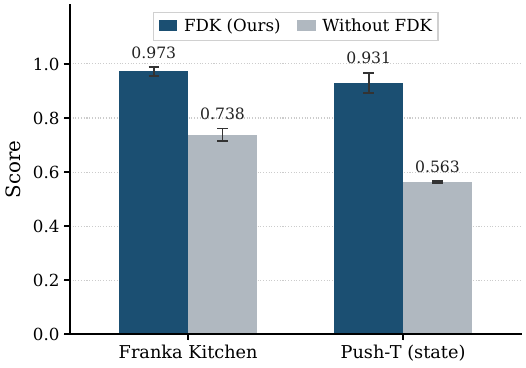}
    \vspace{-6mm}
    \caption{FDK component ablation on Franka Kitchen and state-based Push-T.}
    \label{fig:fdk_component_ablation}  \vspace{-3mm}
\end{wrapfigure}

\textbf{FDK Ablation:} To assess the contribution of the proposed FDK transition, we compare
DNK with a matched variant in which the learned latent transition is
replaced by the identity map, $\mathbf z^{+}=\mathbf z$.
As shown in Fig.~\ref{fig:fdk_component_ablation}, removing FDK reduces
the mean score from $0.973$ to $0.738$ on Franka Kitchen and from
$0.931$ to $0.563$ on state-based Push-T.
These results show that the learned FDK transition contributes
substantially to closed-loop performance across both long-horizon and
contact-rich manipulation tasks.
Full experimental details and qualitative comparisons are provided in
Appendix~\ref{app:fdk_ablation}.\looseness=-1

\begin{wrapfigure}{r}{0.28\textwidth}
    \vspace{-3mm}
    \centering
    \captionsetup{font=small}
    \includegraphics[width=0.98\linewidth]
    {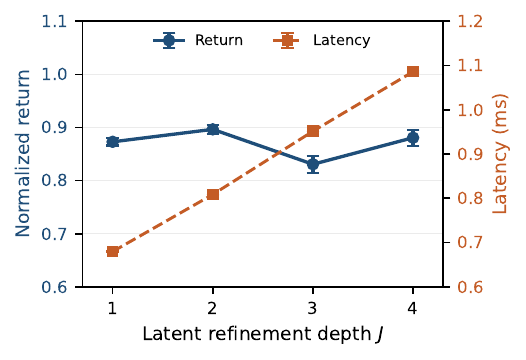}
    \vspace{-6mm}
    \caption{HalfCheetah ablation on latent refinement depth \(J\): normalized return (left axis) and inference latency (right axis).}
    \label{fig:ablation_num_iters}  \vspace{-3mm}
\end{wrapfigure}

\textbf{Ablation on Internal Latent Refinement Depth:}  
To investigate the effect of internal latent refinement, we recursively apply the FDK transition \(J\) times while keeping all other hyperparameters fixed on \texttt{halfcheetah-medium-expert-v2}. This ablation changes only the internal depth of the one-step student.
Figure~\ref{fig:ablation_num_iters} illustrates the return and latency for \(J\in\{1,2,3,4\}\). Latency grows nearly linearly with \(J\) (from \(0.680\,\mathrm{ms}\) at \(J=1\) to \(1.086\,\mathrm{ms}\) at \(J=4\)), while return stays within a band near \(0.83\)--\(0.90\). The highest mean return is at \(J=2\) (\(0.8966 \pm 0.0083\)), but the gain over \(J=1\) (\(0.8731 \pm 0.0070\)) is marginal relative to the added cost.  {This behavior is consistent with our Koopman-inspired evolution: repeated
application of the same linear operator does not introduce new latent
modes, but only rescales the modal gains. Hence, increasing $J$ provides
little benefit while adding computation. We therefore use $J=1$ by
default. }

% \textbf{Ablation on Internal Latent Refinement Depth:}  
% To investigate the effect of internal latent refinement, we recursively apply the FDK transition \(J\) times while keeping all other hyperparameters fixed on \texttt{walker2d-medium-expert-v2}. This ablation changes only the internal depth of the one-step student.
% Figure~\ref{fig:ablation_num_iters} illustrate the return and latency for \(J\in\{1,2,3,4\}\). Latency grows nearly linearly with \(J\) (from \(2.84\,\mathrm{ms}\) at \(J=1\) to \(6.39\,\mathrm{ms}\) at \(J=4\)), while return stays within a narrow band near \(1.095\)--\(1.099\). The highest mean return is at \(J=4\) (\(1.0990 \pm 0.0006\)), but the gain over \(J=1\) (\(1.0963 \pm 0.0005\)) is marginal relative to the added cost. \textcolor{blue}{ This behavior is consistent with our Koopman-inspired evolution: repeated
% application of the same linear operator does not introduce new latent
% modes, but only rescales the modal gains. Hence, increasing $J$ provides
% little benefit while adding computation. We therefore use $J=1$ by
% default. } 
% % We therefore use a single latent transition step (\(J=1\)) as the default setting, since it offers a favorable overall latency--return trade-off with minimal additional computation.

\subsection{Real-World Experiment} 
\label{subsec:real-world} 
\begin{figure}[tp]
    \centering
    \includegraphics[width=0.98\linewidth]{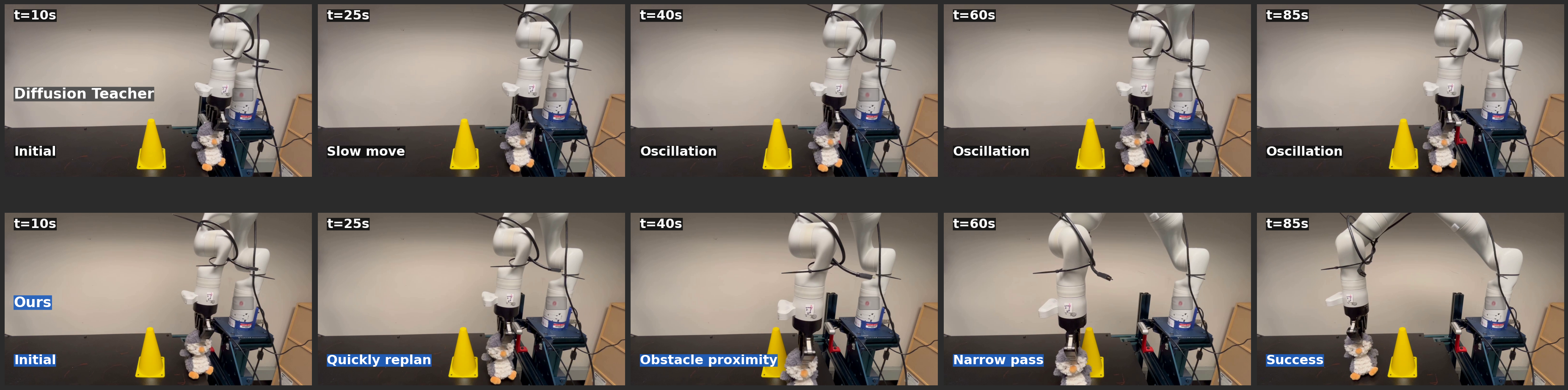}
    \vspace{-0mm}
\caption{Consecutive frames from the real-world Kinova experiment.
The diffusion teacher exhibits repeated corrective motion near the
obstacle, consistent with the mismatch between its slower replanning
rate and the faster low-level control loop.  In contrast, DNK replans more frequently and traverses the narrow region with shorter goal-reaching time.}
    \label{fig:kinova_trajs}
    \vspace{-0mm}
    \end{figure}

To evaluate the proposed method in a real robotic setting, we consider an obstacle-aware reaching task on a Kinova Gen3 manipulator, as shown in Fig.~\ref{fig:kinova_trajs}. The task requires the robot to move its end-effector from an initial position to a goal position while avoiding static obstacles in the workspace. 

Figure~\ref{fig:kinova_trajs} shows a representative rollout in which the diffusion teacher progresses slowly and exhibits repeated corrective motion near the obstacle, whereas DNK replans faster and traverses the narrow passage to reach the goal.
Across 50 hardware trials, DNK reduces mean inference latency from
$151.00$\,ms to $4.08$\,ms and mean task completion time from
$314.95$\,s to $85.43$\,s relative to the diffusion teacher.
Detailed hardware settings and results are provided in
Appendix~\ref{app:hardware_experiments}.
 
 % During execution, the planner operates online in a receding-horizon manner using the current robot state, goal specification, and obstacle representation. 
 % Details of the hardware setup, evaluation protocol, and quantitative results are provided in Appendix~\ref{app:hardware_experiments}.

\section{Conclusion and Future Work}\vspace{-0mm}
\label{sec:con}
This work presents a DNK distillation framework that distills a multistep diffusion policy into a one-step policy through an FDK layer with spectral regularization. The method achieves strong one-step performance across locomotion and manipulation benchmarks and on a real-world Kinova manipulator task,  while substantially reducing inference latency and using a compact student architecture. Experiments on a physical Kinova Gen3 further demonstrate millisecond-scale inference and obstacle-aware reaching under receding-horizon control. The current framework does not explicitly enforce safety constraints or address robustness under broader distribution shifts. Future work will incorporate safety-critical constraints and investigate adaptation to more complex and unstructured robotic environments, such as human-robot interaction scenarios.

\subsection*{AI Use Statement}
Generative AI tools were used for translation, language editing,
LaTeX formatting, and feedback on the technical presentation of the
manuscript. The authors reviewed and verified all AI-assisted revisions,
including mathematical statements and interpretations of experimental
results, and take full responsibility for the final content.

\subsection*{Ethics Statement}
This work involves simulation and controlled laboratory experiments on
robotic systems and does not involve human participants or personal
data. The proposed method does not provide formal safety guarantees;
deployment in safety-critical settings would require application-specific
validation and appropriate safety mechanisms.

% \subsection*{Reproducibility Statement}
% The problem formulation and proposed DNK framework are described in
% Sections~\ref{sec:problem}--\ref{sec:method}.
% Architectural details, training objectives, and implementation settings
% are provided in the corresponding method and experimental appendices.
% Benchmark configurations, hyperparameters, evaluation protocols, and
% hardware settings are documented in Appendix~\ref{app:experimental_setup}
% and Appendix~\ref{app:hardware_experiments}.
% Proofs of the theoretical results are provided in
% Appendix~\ref{app:technical_appendices}.
% We plan to release the source code, trained checkpoints, and evaluation
% scripts under an open-source license upon acceptance.

% We will release anonymized code, checkpoints, calibration details, seeds, and matched-compute scripts. An LLM assisted language editing; the authors verified and remain responsible for all content.

\bibliography{references}
\bibliographystyle{plainnat} % Replace with iclr2026_conference in the official bundle.

% Add appendix by pzq, please add supplementary blow:
\appendix 
\section{Theoretical Properties of the FDK Parameterization}
\label{app:technical_appendices} 
% =========================================================
\subsection{Conditional versus Constant Modal Gains}
\label{app:conditional_gain_proof}
To isolate the effect of the modal-gain parameterization, we analyze context-dependent modal gains with the encoder and shared modal basis $\mathbf V$ in Section~\ref{subsec:fdk_arch} fixed. Let $\mathbf C$ denote the random conditioning variable, with realization $\mathbf c$, and let expectations be taken over the distillation data distribution
\begin{equation}
\mathbf B(\mathbf z^{(1)})
=
\mathbf V
\operatorname{diag}\!\left(
\mathbf V^{-1}\mathbf z^{(1)}
\right),
\label{eq:app_modal_design}
\end{equation}
so that, for a measurable gain function
$g:\mathcal C\rightarrow\mathbb R^L$,
\[
\mathbf B(\mathbf z^{(1)})g(\mathbf c)
=
\mathbf V
\operatorname{diag}\!\bigl(g(\mathbf c)\bigr)
\mathbf V^{-1}\mathbf z^{(1)}.
\]
The FDK transition corresponds to
$g(\mathbf c)=\boldsymbol{\gamma}_\omega(\mathbf c)$.
We consider the population latent-prediction risk
\begin{equation}
R(g)
=
\mathbb E\!\left[
\left\|
\mathcal E_\phi(\tilde{\boldsymbol{\tau}}_0)
-
\mathbf B(\mathbf z^{(1)})g(\mathbf C)
\right\|_2^2
\right].
\label{eq:modal_population_risk}
\end{equation}
Let
\begin{equation}
\begin{aligned}
\mathbf M(\mathbf c)
&=
\mathbb E\!\left[
\mathbf B^\top\mathbf B
\mid \mathbf C=\mathbf c
\right],\\
\mathbf b(\mathbf c)
&=
\mathbb E\!\left[
\mathbf B^\top
\mathcal E_\phi(\tilde{\boldsymbol{\tau}}_0)
\mid \mathbf C=\mathbf c
\right],
\end{aligned}
\label{eq:app_modal_statistics}
\end{equation}
where $\mathbf B=\mathbf B(\mathbf z^{(1)})$.

\begin{proposition}[Conditional versus constant modal gains]
\label{prop:conditional_modal_gains}
Assume the relevant second moments are finite and
$\mathbf M(\mathbf C)\succeq\mu\mathbf I$ almost surely for some
$\mu>0$. Then
\begin{equation}
g^\star(\mathbf c)
=
\mathbf M(\mathbf c)^{-1}\mathbf b(\mathbf c)
\label{eq:app_optimal_conditional_gain}
\end{equation}
minimizes $R(g)$ over admissible conditional gains, and
\begin{equation}
R(g)-R(g^\star)
=
\mathbb E\!\left[
\|g(\mathbf C)-g^\star(\mathbf C)\|_{\mathbf M(\mathbf C)}^2
\right],
\label{eq:modal_excess_risk}
\end{equation}
where $\|\mathbf u\|_{\mathbf M}^2=\mathbf u^\top\mathbf M\mathbf u$.

If $g$ is restricted to be constant, its optimum is
\begin{equation}
g_0
=
\bigl(\mathbb E[\mathbf M(\mathbf C)]\bigr)^{-1}
\mathbb E[\mathbf b(\mathbf C)],
\label{eq:app_optimal_constant_gain}
\end{equation}
and
\begin{equation}
R(g_0)-R(g^\star)
\ge
\mu\,
\mathbb E
\left\|
g^\star(\mathbf C)
-
\mathbb E[g^\star(\mathbf C)]
\right\|_2^2.
\label{eq:conditional_gain_advantage}
\end{equation}
% Thus, conditional gains attain strictly lower optimal risk whenever
% $g^\star(\mathbf C)$ is not almost surely constant.
Thus, the unrestricted conditional-gain class has strictly lower
optimal population risk than the constant-gain class whenever
$g^\star(\mathbf C)$ is not almost surely constant.
\end{proposition}

\begin{proof}
Conditioned on $\mathbf C=\mathbf c$, the risk for
$\mathbf a\in\mathbb R^L$ is
\[
q_{\mathbf c}(\mathbf a)
=
\mathbb E\!\left[
\left\|
\mathcal E_\phi(\tilde{\boldsymbol{\tau}}_0)
\right\|_2^2
\mid \mathbf C=\mathbf c
\right]
-2\mathbf a^\top\mathbf b(\mathbf c)
+\mathbf a^\top\mathbf M(\mathbf c)\mathbf a.
\]
Since $\mathbf M(\mathbf c)\succeq\mu\mathbf I$, this quadratic is
strictly convex, with unique minimizer
$g^\star(\mathbf c)=\mathbf M(\mathbf c)^{-1}\mathbf b(\mathbf c)$.
Completing the square gives
\[
q_{\mathbf c}(\mathbf a)-q_{\mathbf c}(g^\star(\mathbf c))
=
\|\mathbf a-g^\star(\mathbf c)\|_{\mathbf M(\mathbf c)}^2,
\]
and taking expectations yields~\eqref{eq:modal_excess_risk}.

For constant $\mathbf a$,
\[
R(\mathbf a)
=
\mathbb E
\left\|
\mathcal E_\phi(\tilde{\boldsymbol{\tau}}_0)
\right\|_2^2
-2\mathbf a^\top\mathbb E[\mathbf b(\mathbf C)]
+\mathbf a^\top\mathbb E[\mathbf M(\mathbf C)]\mathbf a,
\]
whose unique minimizer is~\eqref{eq:app_optimal_constant_gain}.
Applying~\eqref{eq:modal_excess_risk} to
$g(\mathbf C)\equiv g_0$ gives
\[
R(g_0)-R(g^\star)
\ge
\mu\,\mathbb E
\|g_0-g^\star(\mathbf C)\|_2^2.
\]
Using
\[
\mathbb E\|g_0-g^\star(\mathbf C)\|_2^2
=
\mathbb E
\left\|
g^\star(\mathbf C)-\mathbb E[g^\star(\mathbf C)]
\right\|_2^2
+
\left\|
g_0-\mathbb E[g^\star(\mathbf C)]
\right\|_2^2
\]
proves~\eqref{eq:conditional_gain_advantage}.
\end{proof}
 
The condition $\mathbf M(\mathbf C)\succeq\mu\mathbf I$ requires
sufficient excitation of the modal coordinates. The result applies to
a fixed latent representation and shared modal basis. 

\subsection{Spectral Properties of the FDK Operator}
\label{app:fdk_spectral} 

\begin{proposition}[Spectral structure of the FDK operator]
\label{prop:fdk_spectral} 
% Let $\mathbf V\in\mathbb R^{L\times L}$ be invertible.
For any fixed context $\mathbf c$,   the operator
$\mathbf K_{\omega}(\mathbf c)$ in
\eqref{eq:fdk_operator} is diagonalizable, with eigenvalues
\begin{equation}
\lambda_{\ell}\!\left(\mathbf K_{\omega}(\mathbf c)\right)
=
\gamma_{\omega,\ell}(\mathbf c),
\qquad
\ell=1,\ldots,L.
\end{equation}
Consequently,
\begin{equation}
\rho\!\left(\mathbf K_{\omega}(\mathbf c)\right)
=
\max_{\ell}
\left|
\gamma_{\omega,\ell}(\mathbf c)
\right|.
\label{eq:app_spectral_radius}
\end{equation}
Moreover,
\begin{equation}
\left\|
\mathbf K_{\omega}(\mathbf c)
\right\|_2
\le
\kappa_2(\mathbf V)
\max_{\ell}
\left|
\gamma_{\omega,\ell}(\mathbf c)
\right|,
\label{eq:app_operator_bound}
\end{equation}
where
\[
\kappa_2(\mathbf V)
=
\|\mathbf V\|_2\|\mathbf V^{-1}\|_2
\]
is the spectral condition number of the shared modal basis.
\end{proposition}

\begin{proof}
Let
\[
\mathbf D_{\omega}(\mathbf c)
=
\operatorname{diag}
\bigl(\boldsymbol{\gamma}_{\omega}(\mathbf c)\bigr).
\]
Since
\[
\mathbf K_{\omega}(\mathbf c)
=
\mathbf V
\mathbf D_{\omega}(\mathbf c)
\mathbf V^{-1},
\]
$\mathbf K_{\omega}(\mathbf c)$ is similar to
$\mathbf D_{\omega}(\mathbf c)$ and therefore has the same
eigenvalues, which gives
\eqref{eq:app_spectral_radius}. By submultiplicativity of the induced
$2$-norm,
\[
\left\|
\mathbf K_{\omega}(\mathbf c)
\right\|_2
\le
\|\mathbf V\|_2
\left\|
\mathbf D_{\omega}(\mathbf c)
\right\|_2
\|\mathbf V^{-1}\|_2.
\]
Because
\[
\left\|
\mathbf D_{\omega}(\mathbf c)
\right\|_2
=
\max_{\ell}
\left|
\gamma_{\omega,\ell}(\mathbf c)
\right|,
\]
\eqref{eq:app_operator_bound} follows.
\end{proof}

\begin{corollary}[Fixed-context output sensitivity]
\label{cor:fdk_output_sensitivity}
Fix a context $\mathbf c$ and a latent domain $\mathcal Z$.
Suppose that the decoder $\mathcal D_\psi$ is
$L_{\mathcal D}$-Lipschitz on
$\mathbf K_\omega(\mathbf c)\mathcal Z$.
Then, for any $\mathbf z,\mathbf z'\in\mathcal Z$,
\begin{equation}
\begin{aligned}
&\left\|
\mathcal D_\psi\!\left(\mathbf K_\omega(\mathbf c)\mathbf z\right)
-
\mathcal D_\psi\!\left(\mathbf K_\omega(\mathbf c)\mathbf z'\right)
\right\|_2 \\
&\qquad\le
L_{\mathcal D}\,
\kappa_2(\mathbf V)\,
\|\boldsymbol\gamma_\omega(\mathbf c)\|_\infty\,
\|\mathbf z-\mathbf z'\|_2.
\end{aligned}
\label{eq:fdk_output_sensitivity}
\end{equation}
\end{corollary}

\begin{proof}
The decoder Lipschitz condition gives an upper bound of
$L_{\mathcal D}
\|\mathbf K_\omega(\mathbf c)(\mathbf z-\mathbf z')\|_2$.
Applying the induced-norm inequality and
\eqref{eq:app_operator_bound} yields the result.
\end{proof}

Under the exact similarity parameterization,
$\mathcal L_{\mathrm{spec}}$ directly penalizes modal eigenvalue
magnitudes exceeding unity.
Moreover, for an $L_{\mathcal D}$-Lipschitz decoder, the fixed-context
map
$\mathbf z\mapsto
\mathcal D_\psi(\mathbf K_\omega(\mathbf c)\mathbf z)$
has Lipschitz constant on $\mathcal Z$ bounded by
\[
L_{\mathcal D}\,\kappa_2(\mathbf V)\,
\|\boldsymbol\gamma_\omega(\mathbf c)\|_\infty.
\]
Hence, the spectral penalty discourages large modal amplification,
while the conditioning of $\mathbf V$ and the decoder sensitivity
also determine perturbation amplification at the output.
% This bound characterizes one-step sensitivity and does not by itself
% imply contraction or closed-loop stability.

% \paragraph{Implications for latent sensitivity and regularization.}
% Proposition~\ref{prop:fdk_spectral} identifies the modal gains
% as the eigenvalues of the latent operator and separates their
% contribution from that of the shared basis conditioning.
% Corollary~\ref{cor:fdk_output_sensitivity} further relates these
% quantities to the sensitivity of the decoded output at a fixed
% context.

% The regularizer is nevertheless a soft, expectation-based
% penalty: it does not enforce a uniform bound on the gains,
% nor does it constrain $\kappa_2(\mathbf V)$ or the decoder
% Lipschitz constant.
% Even if all modal eigenvalue magnitudes are at most one,
% a non-orthogonal basis can permit Euclidean amplification.
% The results therefore characterize the latent denoising map,
% rather than establish closed-loop stability.
\section{Experimental Setup}
\label{app:experimental_setup}  
\subsection{OneDP-S Implementation}
\label{app:onedp}

\paragraph{Formulation.}
We adapt the stochastic OneDP-S objective of
\citet{wang2025onestep} to our pretrained CleanDiffuser DDPM teachers. The generator and auxiliary score network adopt the same architecture as the teacher and are initialized as separate copies of the EMA weights from the teacher, while the teacher remains frozen. 
Each iteration first updates the auxiliary network with an unweighted
noise-prediction MSE on detached generator samples, then updates the generator
with a detached score-difference surrogate using $w(t)=\sigma_t^2$.
Gradients through the forward-noised generator output are retained.
Distillation uses observation contexts only and does not regress to
demonstration actions or cached teacher endpoints.

\paragraph{Teacher adaptation.}
We retain each teacher's native diffusion schedule
and normalization statistics.
For Push-T and Franka Kitchen, the generator receives
standard Gaussian noise and converts its noise prediction
at a fixed initialization index $t_{\mathrm{init}}$
into a clean action sequence using the corresponding
diffusion coefficients.

For locomotion, the inherited clean-sample predictor
directly generates a state--action trajectory.
The current normalized state is inpainted at the first
trajectory step, and these fixed entries are excluded
from both training losses.
Teacher and auxiliary clean predictions are converted
to noise predictions for loss computation.
Table~\ref{tab:onedp_settings} summarizes the native
diffusion grids; score-training indices are sampled
uniformly. 

\begin{table}[t]
\centering
\caption{Task-specific OneDP-S settings.
Indices are zero-based.
Each alternating iteration comprises one auxiliary and one generator
optimizer step.}
\label{tab:onedp_settings}
\small
\setlength{\tabcolsep}{5pt}
\begin{tabular}{@{}lcccc@{}}
\toprule
Task & Teacher levels & $t_{\mathrm{init}}$
& Score indices & Iterations \\
\midrule
HalfCheetah / Walker2d
& 20 & 12 & $1$--$18$ & 156{,}260 \\
Push-T
& 5 & 2 & $0$--$3$ & 6{,}840 \\
Franka-Kitchen
& 50 & 32 & $1$--$47$ & 37{,}580 \\
\bottomrule
\end{tabular}\vspace{-2mm}
\end{table}

\paragraph{Optimization.}
All students are trained for 20 epochs with batch size 64 and Adam
with $(\beta_1,\beta_2)=(0,0.999)$.
% The generator and auxiliary learning rates 
The learning rates for the generator and auxiliary network  are $10^{-6}$ and
$2\times10^{-5}$, respectively, with gradient-norm  clipped at 10.
Locomotion training uses 500{,}000 observation contexts per task and
retains frozen reward-classifier guidance with weights $10^{-4}$
(HalfCheetah) and $10^{-3}$ (Walker2d).
In the locomotion implementation, latent noise, diffusion indices, and
forward noise are resampled independently for the auxiliary and generator
updates while sharing the same observation minibatch.
In the manipulation implementation, these random draws are reused across
the two updates.

\paragraph{Evaluation.}
For each task, we evaluate using 3 seeds, each one consisting of 50 episodes.
Locomotion uses trajectory horizons of 4 (HalfCheetah) and 32 (Walker2d).
 A single batched generator evaluation produces 64 candidates, whose first actions are clipped to $[-1,1]$ and ranked by the shared frozen IQL scorer. The selected action is then executed before replanning. 
Manipulation policies condition on two observation frames, predict 16
actions, and execute eight, using a single candidate without reranking.
Image-based Push-T uses RGB observations and pusher position without
privileged object state; its visual encoders remain trainable during
distillation, with deterministic preprocessing.
At inference time we deploy the generator and its conditioning components,
together with the IQL scorer for locomotion.

\paragraph{Inverse implementation.}
The shared basis $\mathbf V$ is a trainable dense matrix initialized
orthogonally. During training, we compute $\mathbf V^{-1}\mathbf z$
by solving $\mathbf V\mathbf u=\mathbf z$ with
\texttt{torch.linalg.solve}, rather than learning an independent
inverse. For frozen inference, $\mathbf V^{-1}$ is computed once using
\texttt{torch.linalg.inv} and cached. The basis is otherwise
unconstrained, and the spectral results assume nonsingularity.

Numerical checks of selected checkpoints using the exact similarity
parameterization (CPU FP32) are reported in
Table~\ref{tab:inverse_residuals}, where the normalized two-sided
inverse residual is defined as
\begin{equation}
r_{\mathrm{inv}}
=
\frac{
\max\!\left\{
\|\mathbf V\widehat{\mathbf V^{-1}}-\mathbf I\|_F,\,
\|\widehat{\mathbf V^{-1}}\mathbf V-\mathbf I\|_F
\right\}}
{\sqrt{L}},
\end{equation}
where $\widehat{\mathbf V^{-1}}$ denotes the numerically computed
cached inverse. The quantity $\delta_{\mathrm{sd}}$ denotes the relative
Frobenius discrepancy between solve-based and cached-inverse latent
outputs over 256 synthetic test inputs.
These diagnostics confirm numerically accurate inverse application and
consistent FDK operator computation between training and deployment.

\begin{table}[t]
\centering
\caption{Numerical diagnostics of the FDK inverse implementation.}
\label{tab:inverse_residuals}
\setlength{\tabcolsep}{4pt}
\begin{tabular}{@{}lcc@{}}
\toprule
Checkpoint & $r_{\mathrm{inv}}$ & $\delta_{\mathrm{sd}}$ \\
\midrule
Push-T (state) & $2.00\times10^{-6}$ & $2.37\times10^{-6}$ \\
Push-T (image) & $2.23\times10^{-6}$ & $2.58\times10^{-6}$ \\
Franka Kitchen & $1.71\times10^{-6}$ & $2.02\times10^{-6}$ \\
Walker2d & $7.31\times10^{-6}$ & $8.84\times10^{-6}$ \\
HalfCheetah & $1.14\times10^{-5}$ & $9.05\times10^{-6}$ \\
\bottomrule
\end{tabular}
\end{table}
\subsection{Deployed Parameter Counts Across Baselines}
\label{app:param_counts}
\paragraph{Code provenance.}
We use the public CleanDiffuser implementation for the DDPM teachers, CT,
and CD,\footnote{\url{https://github.com/CleanDiffuserTeam/CleanDiffuser}}
and the official Consistency Policy codebase for CP and its EDM
teacher.\footnote{\url{https://github.com/Aaditya-Prasad/Consistency-Policy}}
Because OneDP-S has no public official release that we could run unchanged,
we reimplement the published stochastic OneDP-S
objective~\citep{wang2025onestep} on our frozen   DDPM teachers,
as detailed in Appendix~\ref{app:onedp}.
KDM and KDM-F follow the open-source repository of
\citet{berman2025onestep}.\footnote{\url{https://github.com/azencot-group/KDM}}

\paragraph{Counting protocol.}
We count each unique parameter required for inference once, including
frozen visual encoders, but excluding training-only teachers, auxiliary
score networks, optimizer states, and additional weight copies used only
during training.
Unless otherwise stated, CD, CT, and OneDP-S retain the generator backbone
of the corresponding local DDPM teacher $T_D$ and therefore share its
deployed parameter count.
% In particular, OneDP-S initializes its generator from the teacher's EMA
% weights and then optimizes it by distillation: the procedure reduces the
% number of sampling evaluations, not the backbone size.
% CP instead uses an EDM-family architecture adapted to each task; its
% parameter count need not exactly match that of its teacher $T_E$.
{In particular, OneDP-S initializes the generator from the EMA weights of the teacher and subsequently optimizes it through distillation. This procedure reduces the number of sampling evaluations while keeping the backbone size unchanged. In contrast, CP uses an EDM-family architecture adapted to each task, and its parameter count does not necessarily match that of the corresponding teacher $T_E$.}

\paragraph{Reported sizes.}
On low-dimensional Push-T and Franka-Kitchen, DDPM-backbone models contain
approximately $69$M parameters, whereas the EDM teacher and CP contain
approximately $248$--$255$M parameters.
On Push-T, switching from state to image observations increases the total
deployed count by approximately $11.50$M for DDPM/OneDP-S models and
$14.70$M for EDM/CP models.
All image-based counts include the visual conditioning modules.

\subsection{Benchmarks and Setup}  
\label{subsec:sim_setup}
We evaluate DNK on D4RL MuJoCo locomotion and two manipulation
benchmarks, Push-T and Franka Kitchen.
Across benchmarks, DNK uses the general training objective in \eqref{eq:total_loss}, with task-specific coefficients,
training budgets, and teacher sampling settings summarized in
Table~\ref{tab:student_hparams}.

\paragraph{MuJoCo locomotion.}   
We use \texttt{halfcheetah-medium-expert-v2} and
\texttt{walker2d-medium-expert-v2} {from D4RL~\citep{fu2020d4rl}, both with 17D observations
and 6D actions.}
Following CleanDiffuser~\citep{dong2024cleandiffuser}, the planning
horizon is~$4$ for HalfCheetah and~$32$ for Walker2d.
The DDPM teacher uses $20$ reverse steps at inference,  whereas 
the EDM teacher uses $18$ Euler sampling steps.
Our method, KDM, and KDM-F are trained on the same teacher-generated
distillation set under a matched configuration with latent size
\(L{=}512\), batch size~$64$, learning rate \(3\times10^{-4}\), and
$200$ epochs.
For each task, we generate $500{,}000$ teacher--target pairs
\((\boldsymbol{\tau}_{\mathrm{prior}},\tilde{\boldsymbol{\tau}}_0)\).
Evaluation uses~$64$ candidates, IQL scoring, and argmax selection,
with~$3$ seeds,~$50$ rollouts per seed,
and a maximum episode length of~$1000$ steps.
The encoder comprises two 1D convolutions
($128$/$256$ channels, kernel size~$5$, ReLU), adaptive average
pooling, and a linear map to \(z\in\mathbb{R}^{L}\). The decoder is a
two-layer MLP with $1024$ hidden units that reconstructs a trajectory.

\paragraph{Push-T.}   
Push-T~\citep{florence2022implicit} requires a circular pusher to
align a T-shaped block with a fixed target pose. We use the public Diffusion Policy demonstration archive~\citep{chi2025diffusion} under the CleanDiffuser setup~\citep{dong2024cleandiffuser}, which contains $206$ episodes and $25{,}650$ frames. We consider both state and image observations, using either a $5$D state or a $96\times96$ RGB image together with the pusher position. 
Actions are $2$D pusher targets.
The episode score is
\(\max_t\operatorname{clip}(c_{i,t}/0.95,0,1)\),
where \(c_{i,t}\) is target-area coverage.  
Episodes last at most~$300$ steps.
The DDPM teacher uses~$5$ reverse steps and the
\texttt{ChiUNet1d} backbone with FiLM observation
conditioning~\citep{perez2018film}.

\paragraph{State-based Push-T.}  \vspace{-1mm}
We distill about \(1.03\times10^{5}\) pairs for up to
\(1.5\times10^{5}\) updates (batch size~$128$).
The student encodes a length-$H$ trajectory that concatenates the
current state with a noisy action sequence, applies a state-conditioned
latent transition, and decodes the result to the same layout.
Two-frame history is injected with gated FiLM,
\(z\leftarrow z+\alpha(z\odot s+b)\),
where \(\alpha{=}\sigma(\alpha_0)\) starts near zero so training begins
close to an identity map. 
Large FiLM scales and shifts are additionally penalized with
weight~$0.001$. 

\paragraph{Image-based Push-T.} 
The image-based student receives two frames of $96\times96$ RGB
observations together with the corresponding pusher positions. 
{We freeze the visual encoder of the teacher, construct a $512$D condition from the features of the two frames, and distill \(1.0\times10^{5}\) pairs over \(1.0\times10^{5}\) updates with a batch size of $256$. 
A projection maps the condition to $128$D. The temporal encoder processes the noisy action chunk together with this projection, while the latent transition uses the full $512$D visual condition. The temporal decoder then predicts the action chunk. 
No FiLM branch is used. The visual history is incorporated only through the condition and the latent transition.}
% We freeze the teacher's visual encoder, form a $512$D condition from
% the two frame features, and distill \(1.0\times10^{5}\) pairs for
% \(1.0\times10^{5}\) updates (batch size~$256$).
% A projection maps the condition to $128$D; a temporal encoder reads the
% noisy action chunk with this projection; the latent transition uses the
% full $512$D visual condition; and a temporal decoder predicts the action
% chunk.
% There is no FiLM branch: visual history enters only through the
% condition and the latent transition.

\paragraph{Franka Kitchen.}
Franka Kitchen~\citep{gupta2019relay} is a long-horizon Franka
manipulation benchmark.
We use \texttt{kitchen-all-v0}, which {contains} seven
subtasks that may be completed in any order.
From the released multitask demonstrations, {our preprocessing
produces} $603$ episodes.
Observations are $60$D and actions are $9$D absolute joint targets.
Policies condition on two frames, predict~$16$ actions, execute eight,
and {then replan}, with a maximum episode length of~$280$ steps.
The DDPM teacher is a CleanDiffuser \texttt{ChiUNet1d}, trained for
\(10^{6}\) updates and evaluated with EMA weights using~$50$ reverse
steps.
We build $100{,}000$ training and $8{,}192$ held-out pairs from the
frozen teacher, with an episode-level train/validation split.
Each pair contains a two-frame history, Gaussian action noise, and a
teacher-generated \(16\times9\) action sequence; zero-filled
future-state slots are not used as targets.

The student is a temporal convolutional encoder--decoder with channel
widths $128$, $192$, and $256$ and \(L{=}128\)
($2.46$M parameters vs.\ $68.92$M for the teacher), with
observation-gated FiLM and observation-dependent modal gains.
We train with Adam for $150{,}000$ updates (batch size~$128$, learning
rate \(3\times10^{-4}\), EMA~$0.9999$); closed-loop validation selects
the EMA checkpoint at $80{,}000$ updates.
Following~\citep{chi2025diffusion,dong2024cleandiffuser}, let \(p_k\)
denote the fraction of rollouts completing at least \(k\) distinct
subtasks.
The primary score is $S=\frac{1}{4}\sum_{k=1}^{4}p_k$, 
equivalently the mean of \(\min(n,4)/4\), where \(n\) is the number of
distinct completed subtasks. 
We distinguish this normalized completion score from \(p_4\), which
measures the fraction of rollouts completing at least four subtasks.
 
Unless otherwise noted, Push-T and Franka Kitchen are evaluated on three evaluation groups of $50$ episodes
each. Notably, for these action-centric manipulation tasks, prediction supervision is retained through $\mathcal L_{\mathrm{act}}$ on the decoded action sequence; Packed trajectories contain zero-filled future-state slots that are not valid teacher targets, and closed-loop performance depends on executable actions rather than full-state reconstruction.  Hence, we set
\(\alpha_{\mathrm{rec}}{=}\alpha_{\mathrm{pred}}{=}0\). 
 
\begin{table}[t]
\centering
\caption{
Student training settings by benchmark.
For locomotion, the action weighting is the diagonal
\(\mathbf{W}_a\), with \(w_1{=}1.5\) and \(w_{k>1}{=}0.6\).
For manipulation, action-loss entries denote
\((\lambda_{\mathrm{chunk}},\lambda_{\mathrm{full}},
\lambda_{\mathrm{first}})\), corresponding to the executed action
window, the full predicted action sequence, and additional emphasis on
the first executed action, respectively. 
}
\label{tab:student_hparams}
\small
\setlength{\tabcolsep}{3.2pt}
\begin{tabular}{@{}lcccc@{}}
\toprule
 & HalfCheetah / Walker2d
 & Push-T (state)
 & Push-T (image)
 & Franka Kitchen \\
\midrule
Latent size \(L\)
  & 512 & 128 & 128 & 128 \\
Batch size
  & 64 & 128 & 256 & 128 \\
Learning rate
  & \(3\times10^{-4}\)
  & \(3\times10^{-4}\)
  & \(3\times10^{-4}\)
  & \(3\times10^{-4}\) \\
EMA rate
  & 0.9999 & 0.9999 & 0.9999 & 0.9999 \\
Train pairs
  & \(5\times10^{5}\)
  & \({\approx}1.03\times10^{5}\)
  & \(1.0\times10^{5}\)
  & \(1.0\times10^{5}\) \\
\midrule
\(\alpha_{\mathrm{rec}}\), \(\alpha_{\mathrm{pred}}\)
  & 0.8,\,1.0 & 0,\,0 & 0,\,0 & 0,\,0 \\
\(\alpha_{\mathrm{lat}}\), \(\alpha_{\mathrm{spec}}\)
  & 0.2,\,0.01
  & 0.15,\,0.01
  & 0.2,\,0.01
  & 0.15,\,0.01 \\
Action weighting
  & \(w_1{=}1.5,\; w_{k>1}{=}0.6\)
  & (1.0,\,0.25,\,0.5)
  & (1.0,\,0.25,\,0)
  & (1.0,\,0.25,\,0) \\
\bottomrule
\end{tabular}\vspace{-0mm}
\end{table}

\section{Real-World Experiments} 
\label{app:hardware_experiments}
\subsection{{Hardware Platform and Setup}}
To evaluate the proposed method in a real robotic setting, we consider an obstacle-aware reaching task on a Kinova Gen3 manipulator, as shown in Fig.~\ref{fig:kinova_trajs}. The robot moves its end-effector from an initial position to a goal position while avoiding static obstacles in the workspace. During execution, the planner operates online in a receding-horizon manner using the current robot state, goal specification, and obstacle representation.

\begin{figure}[tp]
    \centering
    \includegraphics[width=0.98\linewidth]{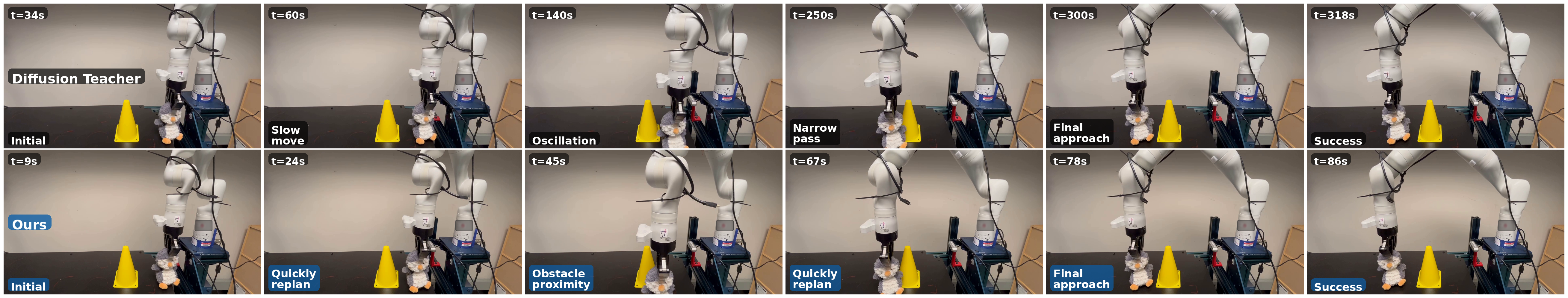}
    \vspace{-0mm}
\caption{Consecutive snapshots of the Kinova manipulator in the real-world setup. The diffusion teacher exhibits slower and repeated
corrective motion near the obstacles, while the proposed student replans
much faster, enabling  more
continuous progress toward the goal.}
    \label{fig:kinova_trajs_appendix}
    \vspace{-4mm}
    \end{figure} 

 The training data are generated offline using a custom-built  B-spline planner over randomized start states, goal states, and obstacle layouts. The resulting collision-aware trajectories are used to train a trajectory diffusion teacher. The proposed student is then trained to distill the teacher policy under the same task conditioning, including the robot state, goal specification, and obstacle representation.  

The teacher and student are evaluated with the same task specification and low-level controller. We use the Kinova Gen3 internal Cartesian position controller, with position targets published at 80 Hz. Each planning request uses the latest available robot state, with at most one request in flight. Once planning completes, only the first action of the highest-scoring candidate is used
to update the Cartesian position target at the next command cycle. Thus, the command publication rate is fixed, whereas the effective replanning rate depends on each
method's computation time. 
During online planning, the DDPM teacher uses 30 reverse diffusion steps. Both methods use a prediction horizon of 32 steps and generate 64 candidates, which are ranked by a geometry-based score that rewards obstacle clearance and
penalizes control effort. Success is defined as the end-effector reaching within
$r_{\mathrm{goal}}=5\,\mathrm{mm}$ of the goal position within $100\,\mathrm{s}$, without end-effector collision with the obstacle during execution. This time threshold is used only to determine success; trials continue beyond $100\,\mathrm{s}$ to measure the full task completion time. 

Each method is evaluated over 50 hardware trials using inference latency, task completion time, minimum obstacle clearance, and success rate. For each trial, we compute the mean and 95th-percentile
inference latency across recorded planning calls, and report each statistic as mean $\pm$ standard deviation across the 50 trials.
% Terminal error is the final end-effector distance to the goal, while 
Minimum end-effector clearance is the minimum distance between the end-effector position and the obstacle surface over the executed trajectory.
All online inference is evaluated on a Titan workstation running Ubuntu 20.04 with two NVIDIA TITAN RTX GPUs, each with 24 GB of VRAM; this specification is reported for reproducibility and does not imply a deployment requirement.

 \begin{table}[t]
\centering
\caption{Hardware comparison on the Kinova reaching task
over 50 trials per method. Continuous metrics are reported as mean $\pm$ standard deviation; success rates are
reported as percentages. Bold denotes the best results}\vspace{-2mm}
\label{tab:kinova_compare}
\renewcommand{\arraystretch}{1.08}
\setlength{\tabcolsep}{4pt}
\scriptsize
\begin{tabular}{lcc}
\toprule
Metric & Diffusion Teacher & Ours \\
\midrule
Mean inference latency (ms) $\downarrow$
& 151.00 $\pm$ 3.73
& \textbf{4.08 $\pm$ 0.20} \\
P95 inference latency (ms) $\downarrow$
& 158.27 $\pm$ 3.78
& \textbf{4.73 $\pm$ 0.23} \\
Completion time (s) $\downarrow$
& 314.95 $\pm$ 3.54
& \textbf{85.43 $\pm$ 2.53} \\
Terminal error (mm) $\downarrow$
& \textbf{3.09 $\pm$ 1.42}
& 3.46 $\pm$ 1.75 \\
Minimum clearance (mm) $\uparrow$
& 46.29 $\pm$ 12.91
& \textbf{48.95 $\pm$ 12.12} \\
Success rate (\%) $\uparrow$
& 0.0 
& \textbf{100.0} \\
\bottomrule
\end{tabular}
\vspace{-4mm}
\end{table}

\subsection{{Results}} \vspace{-2mm}
Table~\ref{tab:kinova_compare} reports the hardware results on the Kinova reconfiguration task. The proposed one-step student reduces the mean inference latency from \(151.00\,\mathrm{ms}\) to \(4.08\,\mathrm{ms}\), corresponding to an approximately \(37.0\times\) reduction relative to the diffusion teacher. Its P95 inference latency is \(4.73\,\mathrm{ms}\), which remains well below the 12.5\,ms command period. In contrast, the teacher cannot refresh the planned target within each
12.5\,ms control period, resulting in a substantially lower effective
replanning rate. This latency reduction is reflected in the task completion time, which decreases from \(314.95\,\mathrm{s}\) to \(85.43\,\mathrm{s}\).

DNK maintains comparable terminal accuracy and obstacle clearance while
substantially reducing inference latency. Our method achieves a \(100.0\%\) success rate over 50 hardware trials. The terminal error remains comparable to the teacher, with \(3.46 \pm 1.75\,\mathrm{mm}\) for our method and \(3.09 \pm 1.42\,\mathrm{mm}\) for the teacher. Additionally, the proposed method achieves a slightly larger mean obstacle clearance, indicating that the faster inference does not compromise obstacle-avoidance behavior in this task.

Figure~\ref{fig:kinova_trajs_appendix} further illustrates the difference in execution behavior. The diffusion teacher exhibits slower corrective motion and visible hesitation near the obstacle, whereas the proposed student replans more promptly and produces smoother obstacle-avoidance behavior before reaching the goal. 
These observations are consistent with the measured completion times and indicate that the distilled student is better suited to closed-loop real-time control on the physical system.

These results support the empirical effectiveness of factorized Koopman distillation for real-time hardware deployment.

\begin{table}[t]
\centering
\caption{ 
FDK component ablation on Franka Kitchen and state-based Push-T. Scores and latencies are reported as mean $\pm$ standard deviation
across three  evaluation groups of 50 episodes each.
Every method uses one network evaluation per action chunk.
}
\label{tab:fdk_component_ablation}
\small
\setlength{\tabcolsep}{4pt}
\begin{tabular}{@{}llccc@{}}
\toprule
Task
& Variant
& Params.\ (M)
& Score $\uparrow$
& Latency (ms) $\downarrow$ \\
\midrule
Franka Kitchen
& FDK (Ours)
& 2.457
& $0.973 \pm 0.016$
& $1.485 \pm 0.004$ \\
Franka Kitchen
& Without FDK
& 2.416
& $0.738 \pm 0.023$
& $1.128 \pm 0.003$ \\
\midrule
Push-T (state)
& FDK (Ours)
& 2.367
& $0.931 \pm 0.037$
& $2.2 \pm 1.0$ \\
Push-T (state)
& Without FDK
& 2.333
& $0.563 \pm 0.004$
& $1.7 \pm 0.9$ \\
\bottomrule
\end{tabular}  
\end{table}

\section{Discussions} 

\begin{figure}[tp]
    \centering
    \includegraphics[width=0.98\linewidth]
    {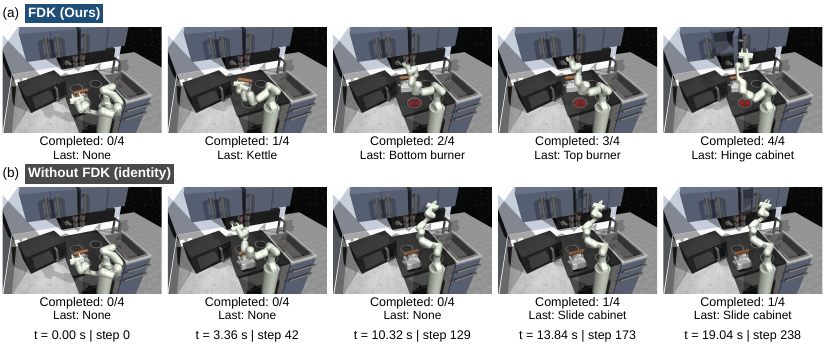}
    \caption{Qualitative FDK component ablation on Franka Kitchen. Time-aligned snapshots compare the full FDK student (top) with the identity variant ($z^{+}=z$, bottom), starting from identical simulator states and initial observations. The FDK student sequentially completes the kettle, bottom-burner, top-burner, and hinge-cabinet subtasks by 19.04\,s. The identity variant completes only the slide-cabinet subtask within the 280-step episode horizon (22.40\,s). Timestamps denote simulated elapsed time rather than wall-clock inference latency. } \vspace{-4mm}
    \label{fig:relay_kitchen_fdk_rollout}
\end{figure}  

\subsection{FDK Component Ablation}
\label{app:fdk_ablation}
To assess the contribution of the FDK transition, we replace
$\mathbf z^{+}=\mathbf V
\operatorname{diag}(\boldsymbol{\gamma}(\mathbf c))
\mathbf V^{-1}\mathbf z$
with the identity map $\mathbf z^{+}=\mathbf z$ on Franka Kitchen and state-based Push-T.
% The remainder of the student---the temporal encoder and decoder,
% observation conditioning, time embedding, and latent normalization---is
% unchanged.
Both variants share the same retained-parameter initialization, data
order, and action and latent-alignment losses. The identity variant
additionally drops the spectral regularizer that applies only to FDK.
Each run uses 150,000 updates and the task-specific checkpoint rule
from Section~\ref{sec:setup}.

\paragraph{Results.}
Table~\ref{tab:fdk_component_ablation} shows that removing FDK sharply
reduces task performance on both benchmarks, despite a smaller model and
slightly lower latency.
{The mean score falls from $0.973$ to $0.738$ on
Franka Kitchen ($-0.235$) and from $0.931$ to $0.563$ on Push-T
($-0.368$), while the parameter count decreases by only about
$1.7\%$ and $1.4\%$, respectively.}
These results show that the learned FDK transition contributes substantially to closed-loop performance relative to an identity latent mapping under the matched training setup. 

Figure~\ref{fig:relay_kitchen_fdk_rollout} compares the two Franka Kitchen
policies  using time-aligned snapshots from the same
initial simulator states and observations.
Ours sequentially completes the   kettle, bottom-burner, top-burner, and hinge-cabinet
subtasks by $19.04$\,s of simulated time.
In contrast, the identity variant completes only the
slide-cabinet subtask within the 280-step episode horizon
(22.40\,s).
This example illustrates the observed performance difference
across successive manipulation stages. 

\subsection{Ablation on Context-Dependent Modal Gains}
\label{app:kitchen_gain_ablation}

To isolate the contribution of context-dependent modal adaptation, we
compare the default FDK transition
\begin{equation}
\mathbf K_{\omega}(\mathbf c)
=
\mathbf V
\operatorname{diag}\!\bigl(
\boldsymbol{\gamma}_{\omega}(\mathbf c)
\bigr)
\mathbf V^{-1} \notag
\end{equation}
against matched static-gain variants on Franka Kitchen.

All variants retain the same encoder, decoder, conditioned trajectory
prior, FiLM conditioning, shared modal basis $\mathbf V$, training data,
and optimization protocol, and differ only in the parameterization of
the modal gains. The \emph{static-vector} variant replaces
$\boldsymbol{\gamma}_{\omega}(\mathbf c)$ with a directly learned,
context-independent vector
$\boldsymbol{\gamma}_{0}\in\mathbb R^{128}$, yielding
$\mathbf V\operatorname{diag}(\boldsymbol{\gamma}_{0})\mathbf V^{-1}$.
To provide a parameter-matched control, the \emph{static-head} variant
retains the same gain-network architecture and parameter count as the
conditional model, but replaces the current observation input with a
fixed training-set reference. Its modal gains are therefore independent
of the current observation while preserving the architecture of the
conditional gain head.

\begin{table}[t]
\centering
\caption{
Matched ablation of modal-gain conditioning on Franka Kitchen.
Scores and latencies are reported as mean~$\pm$~standard deviation.
Scores are computed across three  evaluation groups of $50$
episodes each. Bold denotes the best results.
}
\label{tab:kitchen_gain_ablation}
\small
\setlength{\tabcolsep}{6pt}
\renewcommand{\arraystretch}{1.08}
\begin{tabular}{@{}lccc@{}}
\toprule
Variant
& Params.\ (M)
& Score~$\uparrow$
&Latency~(ms)~$\downarrow$ \\
\midrule
Conditional $\boldsymbol{\gamma}_{\omega}(\mathbf c)$
& 2.457
& $\mathbf{0.973{\pm}0.016}$
& $1.485{\pm}0.004$ \\
Static vector $\boldsymbol{\gamma}_{0}$
& 2.433
& $0.945{\pm}0.018$
& $\mathbf{1.454{\pm}0.007}$ \\
Static head (fixed reference)
& 2.457
& $0.910{\pm}0.005$
& $1.516{\pm}0.014$ \\
\bottomrule
\end{tabular}    
\end{table}

\paragraph{Results.}
Table~\ref{tab:kitchen_gain_ablation} shows that the
context-dependent model achieves a score of
$0.973{\pm}0.016$, compared with $0.945{\pm}0.018$ for the
static-vector variant and $0.910{\pm}0.005$ for the
parameter-matched static-head variant.
Importantly, both static variants retain observation dependence through
the conditioned trajectory prior and FiLM pathways; only the modal gains
are prevented from adapting to the current observation.
The resulting performance gap therefore provides evidence that, on the
long-horizon Kitchen task, these alternative conditioning pathways do
not fully substitute for observation-dependent adaptation of the modal
transition.

The improved closed-loop performance of the conditional-gain model
provides complementary empirical evidence for the benefit of
context-dependent modal adaptation. At the latent level,
Proposition~\ref{prop:conditional_modal_gains} in
Appendix~\ref{app:conditional_gain_proof} shows that, for a fixed
latent representation and shared modal basis, optimal
context-dependent gains can achieve lower latent-prediction risk than
constant gains when the optimal gains vary with context.

\begin{table}[t]
\centering
\caption{
Ablation of the first-step weight in $\mathcal{L}_{\mathrm{act}}$ on
state-based Push-T. Scores are reported as mean $\pm$ standard deviation across three
 evaluation groups of 50 episodes each.
``Worst group'' denotes the minimum group mean, and ``Lower 10\%'' denotes
the mean score over the lowest-performing 10\% of episodes.
Latency follows the matched batch-$1$ host-to-host protocol in
Table~\ref{tab:pusht_robomimic_score_latency}.
}
\label{tab:ablation_lact_first}
\small
\setlength{\tabcolsep}{4pt}
\renewcommand{\arraystretch}{1.08}

\resizebox{\textwidth}{!}{%
\begin{tabular}{@{}lcccccc@{}}
\toprule
Variant
& $\lambda_{\mathrm{first}}$
& Score~$\uparrow$
& Worst group~$\uparrow$
& Lower 10\%~$\uparrow$
&Latency~(ms)~$\downarrow$
& Params.~(M)~$\downarrow$ \\
\midrule
DNK (main)
& $0.5$
& $\mathbf{0.931{\pm}0.037}$
& $\mathbf{0.904}$
& $\mathbf{0.378}$
& $1.283{\pm}0.007$
& $2.37$ \\
DNK, no first-step emphasis
& $0.0$
& $0.913{\pm}0.030$
& $0.887$
& $0.212$
& $1.283{\pm}0.007$
& $2.37$ \\
\bottomrule
\end{tabular}%
}
\end{table}

\subsection{Ablation on Control-Oriented Supervision Loss $\mathcal{L}_{\mathrm{act}}$}
\label{app:ablation_lact_first}
To isolate the effect of emphasizing the first action, we consider the control-oriented action loss in the Push-T task
\begin{equation*}
\mathcal{L}_{\mathrm{act}}
=
\mathcal{L}_{\mathrm{chunk}}
+ \lambda_{\mathrm{full}}\,\mathcal{L}_{\mathrm{full}}
+ \lambda_{\mathrm{first}}\,\mathcal{L}_{\mathrm{first}}.
\end{equation*}
We retain the DNK architecture and training setup and modify only the
first-step weight, setting
$\lambda_{\mathrm{first}}=0.5\rightarrow0$ while keeping
$\lambda_{\mathrm{full}}=0.25$ unchanged on state-based Push-T. 

Table~\ref{tab:ablation_lact_first} reports the results over three
evaluation groups.
Because this ablation changes only the training-time loss weights, the
deployed parameter count and inference latency are identical to
those of the main DNK model in
Table~\ref{tab:pusht_robomimic_score_latency}.
Removing the first-step emphasis reduces the mean score from
$0.931$ to $0.913$ and the worst-group score from $0.904$ to $0.887$.
The degradation is more pronounced in the lower tail, where the
lower-$10\%$ episode mean decreases from $0.378$ to $0.212$.
These results suggest that first-step emphasis provides a modest improvement in average performance in push-T (state) task while more substantially improving performance on difficult episodes, without affecting inference cost. 
This behavior is consistent with chunked receding-horizon control:
each decision executes an eight-step action chunk before replanning, so inaccuracies in the first
executed action propagate through the remainder of the chunk.
% Equivalently, $\lambda_{\mathrm{first}}$ reweights temporal credit in
% $\mathcal{L}_{\mathrm{act}}$ toward the onset of interaction---the step that
% most strongly determines contact initiation and the initial push---while the
% subsequent actions primarily refine an already committed trajectory.
 
\subsection{{Multimodal Candidate Generation}} 
\begin{wrapfigure}{r}{0.58\textwidth}
\vspace{-0.40cm}
\centering
\captionsetup{font=small}
\captionsetup[subfigure]{
    labelfont={small,sf},
    textfont={small,sf}
}
\begin{subfigure}[t]{0.48\linewidth}
    \centering
    \includegraphics[width=\linewidth]
    {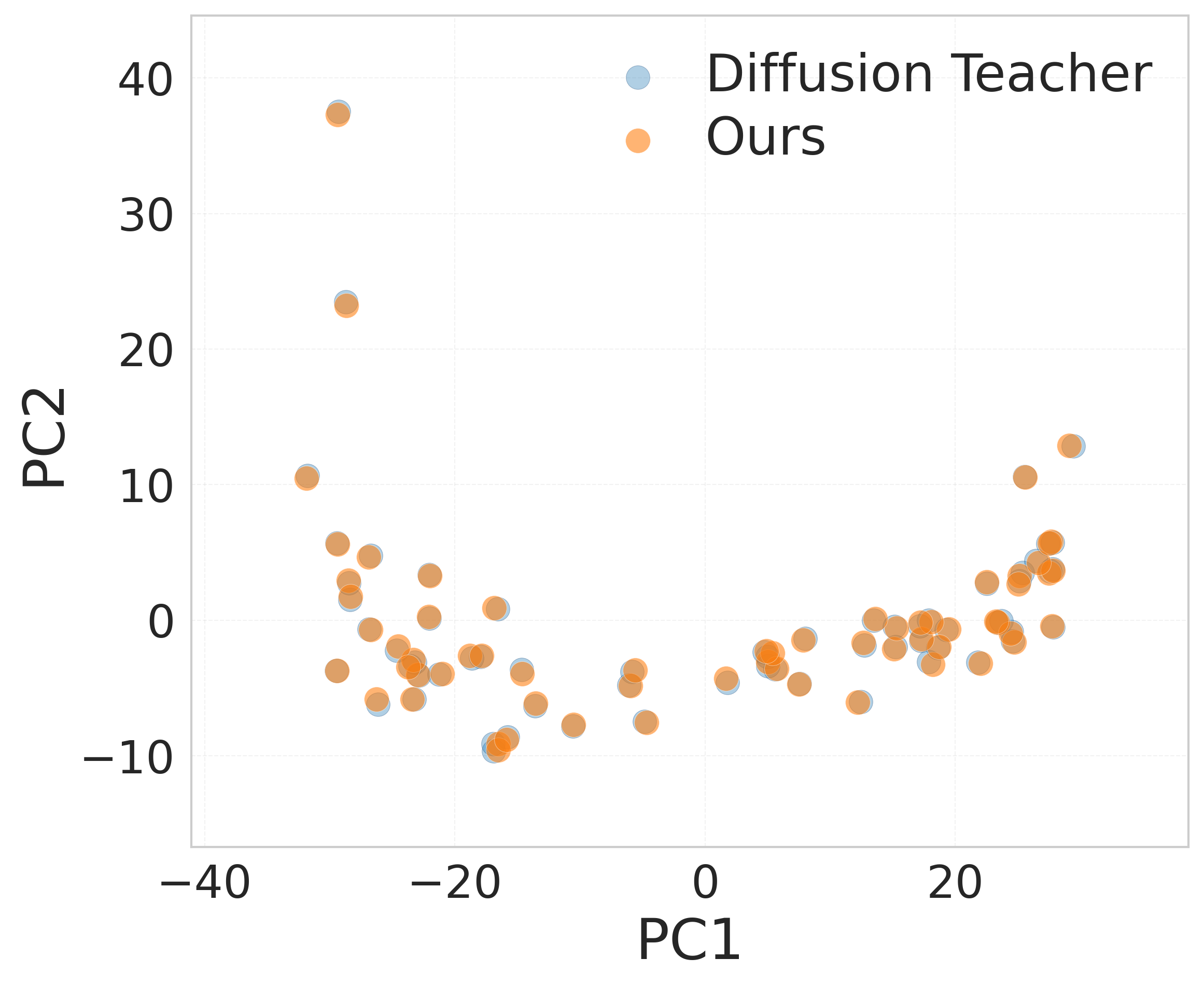}
    \caption{Walker2d.}
    \label{fig:pca_walker2d}
\end{subfigure}
\hfill
\begin{subfigure}[t]{0.48\linewidth}
    \centering
    \includegraphics[width=\linewidth]
    {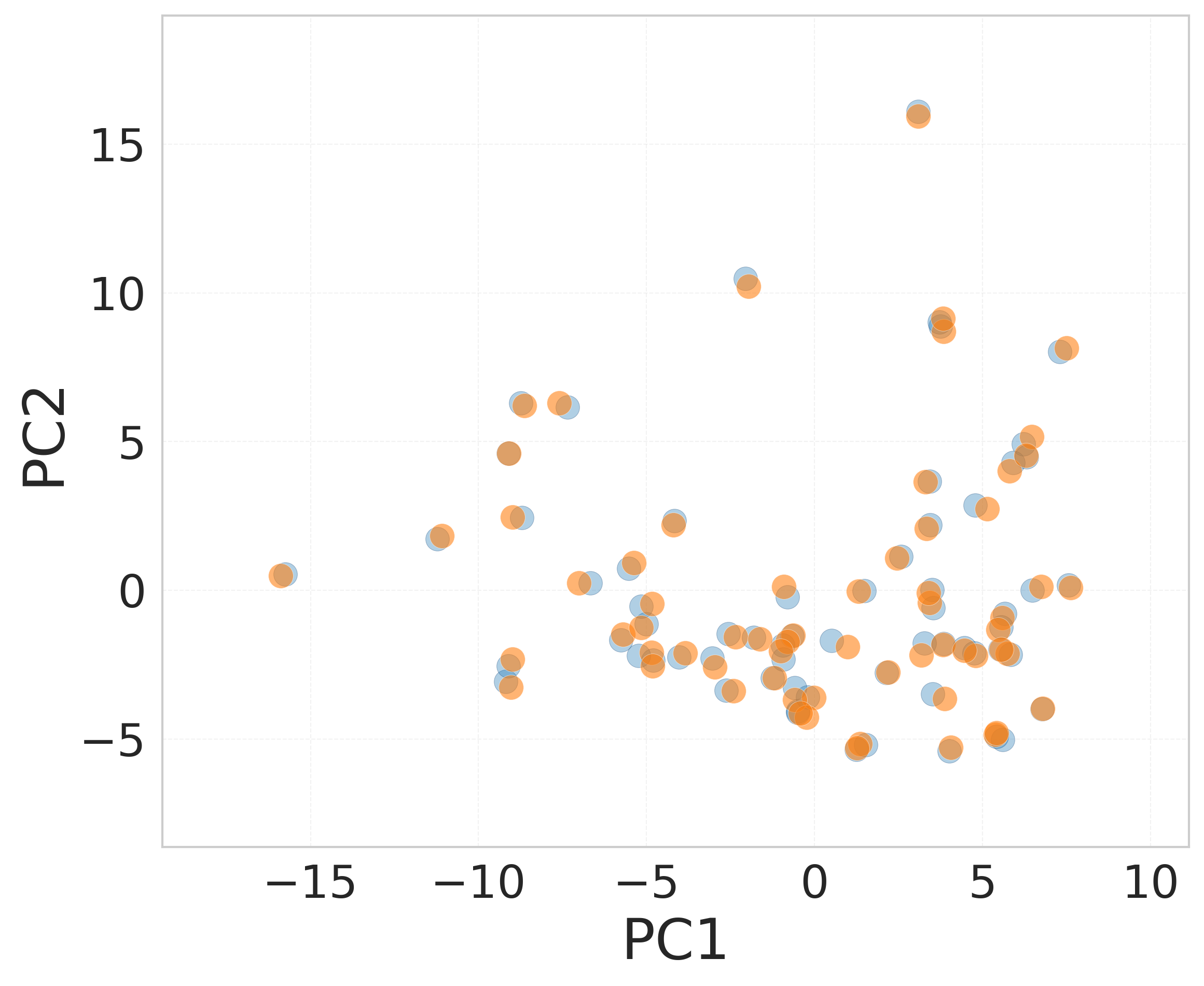}
    \caption{HalfCheetah.}
    \label{fig:pca_halfcheetah}
\end{subfigure}
\vspace{-1mm}
\caption{PCA projection of 64 candidate trajectories generated
from the same initial state. The proposed method preserves broad
candidate coverage with substantial overlap with teacher samples in the projected space.}
\label{fig:pca_multimodality_two_envs}
\vspace{-4mm}
\end{wrapfigure} 
To assess whether the one-step student retains diversity across sampled candidates, we compare trajectories generated by DNK and the diffusion teacher from the same initial state. Fig.~\ref{fig:pca_multimodality_two_envs} shows that the candidate trajectories generated by our method occupy a broad, teacher-aligned region in the principal components. Across both locomotion tasks, DNK produces a broad set of candidates with substantial overlap with the teacher samples in the projected space. Together with the competitive closed-loop returns in Table~\ref{tab:sim_locomotion_results}, this result provides qualitative evidence that the distilled student retains teacher-aligned candidate diversity while using one-step inference.

\subsection{Reduced-Capacity OneDP-S Stress Test}
\label{app:relay_kitchen_capacity}

The one-step baselines differ substantially in deployed model capacity.
On Franka Kitchen, OneDP-S \citep{wang2025onestep} retains the $68.92$M-parameter architecture
of its diffusion teacher, whereas DNK uses only $2.46$M parameters.
To examine whether DNK's performance at this compact scale can be
reproduced by simply narrowing OneDP-S, we evaluate reduced-width
OneDP-S students using the same OneDP-S distillation objective and
frozen full-size DDPM teacher.

All variants use the same observation and action horizons, execute eight
actions per replanning step, use a single policy sample, and are
evaluated on the same three groups of $50$ episodes used in the main
comparison. Changing the U-Net width makes the reduced students
shape-incompatible with the teacher parameters, preventing direct
initialization from the teacher's EMA weights. The reduced-width
variants are therefore trained from random initialization. Accordingly,
we treat this experiment as a capacity stress test of the OneDP-S
training protocol rather than an architecture-matched comparison with
DNK.

\begin{table}[t]
\centering
\small
\renewcommand{\arraystretch}{1.06}
\setlength{\tabcolsep}{4pt}
\caption{Reduced-capacity OneDP-S stress test on Franka Kitchen.
Scores are reported as mean $\pm$ standard deviation across three
 evaluation groups of $50$ episodes each.}
\label{tab:relay_kitchen_capacity}
\begin{tabular}{@{}lcccc@{}}
\toprule
Method & Student width & Params.\ (M) $\downarrow$ & Score $\uparrow$ & Latency (ms) $\downarrow$ \\
\midrule
OneDP-S (full)
& $256$ & $68.92$ & $0.975\pm0.013$ & $6.434\pm0.045$ \\
OneDP-S (reduced)
& $128$ & $19.50$ & $0.055\pm0.009$ & $4.061\pm0.837$ \\
OneDP-S (reduced)
& $64$ & $6.22$ & $0.068\pm0.020$ & $3.953\pm0.789$ \\
OneDP-S (reduced)
& $32$ & $2.44$ & $0.045\pm0.005$ & $3.857\pm0.816$ \\
\bottomrule
\end{tabular}\vspace{-4mm}
\end{table}

\paragraph{Results.}
Table~\ref{tab:relay_kitchen_capacity} shows that reducing the OneDP-S backbone substantially degrades closed-loop
performance. The $19.50$M , $6.22$M, and $2.44$M-parameter variants
achieve scores of $0.055$, $0.068$, and $0.045$, respectively, whereas
DNK achieves $0.973$ with $2.46$M parameters. Notably, even the
$19.50$M OneDP-S variant, which contains nearly eight times as many
parameters as DNK, performs poorly under the reduced-width training
protocol. These results show that simply reducing the OneDP-S backbone to a
comparable parameter scale does not recover DNK's
performance--capacity operating point under the tested training
protocol.

% \subsection{{Performance and parameter size ablation}} 

% \subsubsection{{Real-Time Performance and the Pareto Frontier}} 
% \begin{wrapfigure}[11]{r}{0.50\textwidth}  \centering \captionsetup{font=small} \captionsetup[subfigure]{ labelfont={small,sf}, textfont={small,sf} } \begin{subfigure}[t]{0.48\linewidth} \centering \includegraphics[width=\linewidth] {figures/pareto_walker2d.png} \caption{Walker2d.} \label{fig:pareto_walker2d} \end{subfigure} \hfill \begin{subfigure}[t]{0.48\linewidth} \centering \includegraphics[width=\linewidth] {figures/pareto_halfcheetah.png} \caption{HalfCheetah.} \label{fig:pareto_halfcheetah} \end{subfigure} \vspace{-2mm} \caption{Latency--performance trade-off on D4RL MuJoCo tasks. The x-axis shows mean decision latency on a log scale and the y-axis shows normalized return.} \label{fig:latency_performance_two_envs}  \end{wrapfigure}

% Figure~\ref{fig:latency_performance_two_envs} further illustrates the latency--return trade-off across policy families. Our method achieves a favorable latency--return trade-off, maintaining teacher-level return at low latency. This operating point is desirable for high-frequency closed-loop control, where multistep diffusion policies can be too slow for practical deployment.  

\end{document}